\documentclass[runningheads]{llncs}
\usepackage[T1]{fontenc}
\usepackage{graphicx}
\usepackage{hyperref}
\usepackage{color}

\usepackage[switch]{lineno}
\usepackage{newfloat}
\usepackage{listings}
\usepackage[accsupp]{axessibility}  % % The "axessiblity" package can be found at: https://ctan.org/pkg/axessibility?lang=en. Improves PDF readability for those with disabilities.
\usepackage{epsfig}
\usepackage{amsmath}
\usepackage{amssymb}
\usepackage{wrapfig} % 
\usepackage{glossaries}
\usepackage[table]{xcolor}
\usepackage{colortbl}
\usepackage{bbding}
\usepackage{tikz}
\usepackage{comment}
\usepackage{color}
\usepackage{xspace}
\usepackage{booktabs}
\usepackage{nicefrac}
\usepackage{bm}
\usepackage{tabularx,verbatim}
\usepackage{multirow}
\usepackage{algorithm}
\usepackage{algorithmic}
\usepackage{times}  % DO NOT CHANGE THIS
\usepackage{helvet}  % DO NOT CHANGE THIS
\usepackage{courier}  % DO NOT CHANGE THIS
\usepackage{graphicx} % DO NOT CHANGE THIS

\usepackage{caption} % DO NOT CHANGE THIS AND DO NOT ADD ANY OPTIONS TO IT

\definecolor{cherrypink}{rgb}{0.86,0.29,0.29}
\definecolor{royalblue(web)}{rgb}{0.25, 0.41, 0.88}
\definecolor{darkmagenta}{rgb}{0.55, 0.0, 0.55}

\usepackage[most]{tcolorbox}

\hypersetup{
  breaklinks,
  colorlinks,
}

\usepackage{amsmath,amsfonts,bm}

\def\eqref#1{equation~\ref{#1}}
\def\1{\bm{1}}

\def\vl{{\bm{l}}}

\def\vx{{\bm{x}}}
\def\vy{{\bm{y}}}
\def\vz{{\bm{z}}}

\DeclareMathAlphabet{\mathsfit}{\encodingdefault}{\sfdefault}{m}{sl}
\SetMathAlphabet{\mathsfit}{bold}{\encodingdefault}{\sfdefault}{bx}{n}

\newcommand{\E}{\mathbb{E}}

\newcommand{\R}{\mathbb{R}}

\newcommand{\lesspace}{\vspace{-0.4cm}}
\newcommand{\tablestyle}[2]{\setlength{\tabcolsep}{#1}\renewcommand{\arraystretch}{#2}\centering\footnotesize}

\newcommand{\upperb}{\texttt{Joint (frozen)}\xspace}
\newcommand{\upperbpp}{\texttt{Joint (unfrozen)}\xspace}

\newcommand{\cosnormed}{cosine normalized}
\newcommand{\compcn}{CosNorm\xspace}
\newcommand{\compcnlong}{cosine normalization\xspace}
\newcommand{\compcntitle}{Cosine Normalization\xspace}

\newcommand{\compfr}{KTRFR\xspace}
\newcommand{\compfrlong}{knowledge transfer with robust feature replay\xspace}
\newcommand{\compfrtitle}{Knowledge Transfer with Robust Feature Replay\xspace}
\newcommand{\ncd}{NCD\xspace}

\newcommand{\ncdtitle}{Novel Class Discovery\xspace}

\newcommand{\incd}{iNCD\xspace}

\newcommand{\incdtitle}{Incremental Novel Class Discovery\xspace}

\newcommand{\cincd}{class-iNCD\xspace}
\newcommand{\cincdlong}{class-incremental novel class discovery\xspace}
\newcommand{\cincdtitle}{Class-incremental Novel Class Discovery\xspace}

\newcommand{\il}{IL\xspace}

\newcommand{\uil}{UIL\xspace}
\newcommand{\uillong}{unsupervised incremental learning\xspace}
\newcommand{\uiltitle}{Unsupervised Incremental Learning\xspace}

\newcommand{\cil}{class-IL\xspace}

\newcommand{\forget}{catastrophic forgetting\xspace}

\newcommand{\vitbsixteen}{ViT-B/16\xspace}

\newcommand{\forgetting}{\mathcal{F}}
\newcommand{\accuracy}{\mathcal{A}}

\newcommand{\sinkhorn}{Sinkhorn-Knopp\xspace}

\newcommand{\ranks}{AutoNovel \xspace}
\newcommand{\ocra}{OCRA \xspace}

\newcommand{\ours}{\texttt{Baseline}\xspace}
\newcommand{\ourspp}{\texttt{Baseline++}\xspace}

\newcommand{\protomean}{\bm{\mu}\xspace}
\newcommand{\protovar}{\bm{v}^{2}\xspace}

\newcommand{\tst}{\ensuremath{^{\mathtt{[t]}}}}
\newcommand{\tstminus}{\ensuremath{^{\mathtt{[t-1]}}}}

\newcommand{\tstminuscap}{\ensuremath{^{\mathtt{[T-1]}}}}

\newcommand{\tstotwo}{\ensuremath{^{\mathtt{[1:2]}}}}

\newcommand{\tstotminus}{\ensuremath{^{\mathtt{[1:t-1]}}}}

\newcommand{\tstot}{\ensuremath{^{\mathtt{[1:t]}}}}
\newcommand{\tstoend}{\ensuremath{^{\mathtt{[1:T]}}}}
\newcommand{\tstoendminusone}{\ensuremath{^{\mathtt{[1:T-1]}}}}

\newcommand{\tsi}{\ensuremath{^{\mathtt{[i]}}}}

\newcommand{\tsend}{\ensuremath{^{\mathtt{[T]}}}}
\newcommand{\tsj}{\ensuremath{^{\mathtt{[j]}}}}

\newcommand{\tsone}{\ensuremath{^{[1]}}}
\newcommand{\tstwo}{\ensuremath{^{[2]}}}
\newcommand{\tsthree}{\ensuremath{^{[3]}}}
\newcommand{\tsfour}{\ensuremath{^{[4]}}}
\newcommand{\tsfive}{\ensuremath{^{[5]}}}

\newcommand{\data}{\mathcal{D}}

\newcommand{\classes}{\mathcal{C}}

\newcommand{\numdatast}{\vert\data\tst\vert}

\newcommand{\task}{\mathcal{T}}
\newcommand{\taskb}{\bm{\mathcal{T}}}

\newcommand{\X}{\mathcal{X}}
\newcommand{\Y}{\mathcal{Y}}

\newcommand{\Mu}{\bm{M}}

\newcommand{\eg}{{e}.{g}.}

\newcommand{\ie}{{i}.{e}.}
\definecolor{darkgreen}{RGB}{119,185,0}
\definecolor{cyan}{rgb}{0.831,0.901,0.945}
\definecolor{remark}{rgb}{1,.5,0} 
\definecolor{citecolor}{rgb}{0,0.443,0.737} 
\definecolor{linkcolor}{rgb}{0.956,0.298,0.235} 

\begin{document}

\title{Large-scale Pre-trained Models are Surprisingly Strong in Incremental Novel Class Discovery}

%%%%%%%%% Authors
\titlerunning{Strong Baselines for Incremental Novel Class Discovery}
\author{
Mingxuan Liu\inst{1} \quad
Subhankar Roy\inst{3} \quad
Zhun Zhong\inst{4}\thanks{Corresponding author: Zhun Zhong.}\quad
Nicu Sebe\inst{1} \quad
Elisa Ricci\inst{1,2}
}
\authorrunning{
% Mingxuan Liu, Subhankar Roy, Zhun Zhong, Nicu Sebe, Elisa Ricci
Liu. et al.
}
% If there are more than two authors, 'et al.' is used. $^{3}$ Hefei University of Technology, Hefei, China \quad

%
\institute{
University of Trento, Trento, Italy\\
\email{mingxuan.liu@unitn.it}\\ \and
Fondazione Bruno Kessler, Trento, Italy \and
University of Aberdeen, Aberdeen, UK \and
University of Nottingham, Nottingham, UK
% Princeton University, Princeton NJ 08544, USA \and
}
%
% \maketitle              % typeset the header of the contribution

% \titlerunning{{Large-scale Pre-trained Models are Surprisingly Strong in class-iNCD}}% Part of RIGHT running header

%%%%%%%%% TEASER
% \twocolumn[{%/
% \vspace{-3em}
\maketitle
\begin{center}
  \centering
   \includegraphics[width=0.99\textwidth]{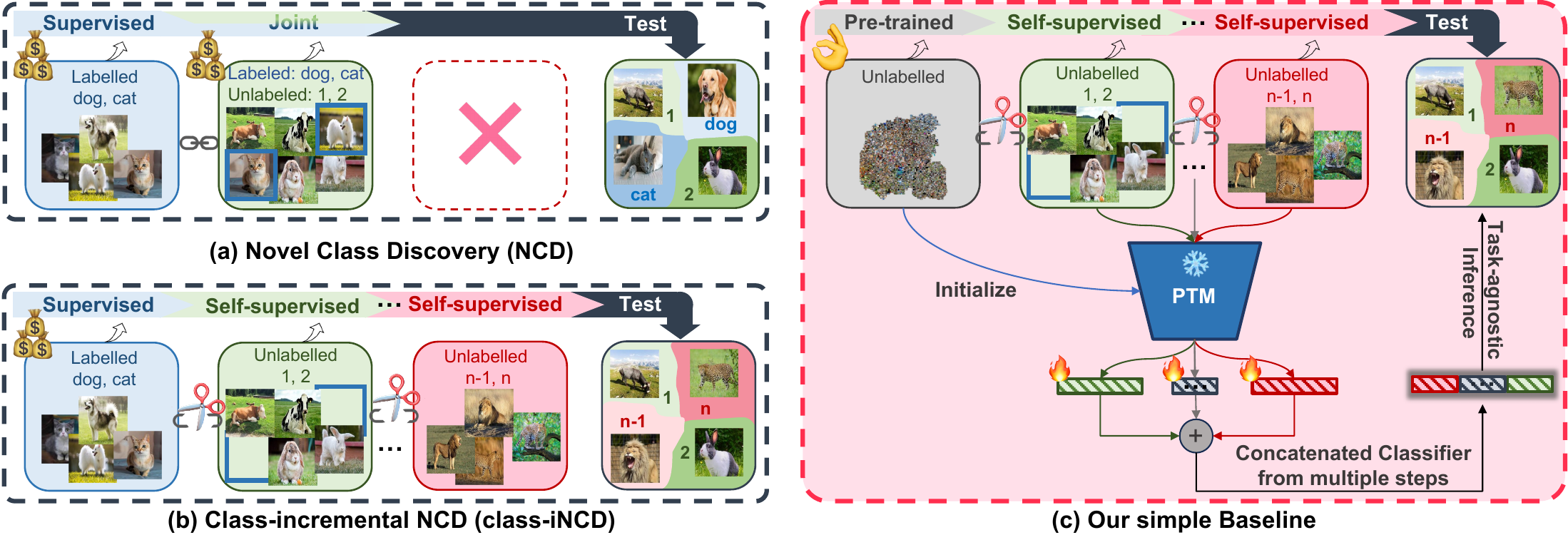}
   \lesspace
   \captionof{figure}{Overview of different learning paradigms for discovering novel (or \textit{new}) categories from \textit{unlabelled} data. (a) \textbf{\ncd} learns and discovers novel classes in an unalabelled dataset by exploiting the priors learned from related labelled data. (b) \textbf{\cincd} is similar to \ncd, except it discovers novel classes arriving in sessions without any access to labelled data during the discovery phase. (c) Our proposed simple Baseline for \cincd that leverages a self-supervised pre-trained model (PTM) instead of expensive labelled data. Inference on test data is carried out in a task-\textit{agnostic} manner.}
   \lesspace
   \label{fig:setting_comparison}
\end{center}
% }]

% \linenumbers

%%%%%%%%% ABSTRACT
\begin{abstract}
   Discovering novel concepts in unlabelled datasets and in a continuous manner is an important desideratum of lifelong learners. In the literature such problems have been partially addressed under very restricted settings, where novel classes are learned by jointly accessing a related labelled set (e.g., \ncd) or by leveraging only a supervisedly pre-trained model (e.g., \cincd). In this work we challenge the status quo in \cincd and propose a learning paradigm where class discovery occurs continuously and truly unsupervisedly, without needing any related labelled set. In detail, we propose to exploit the richer priors from strong self-supervised pre-trained models (PTM). To this end, we propose simple baselines, composed of a frozen PTM backbone and a learnable linear classifier, that are not only simple to implement but also resilient under longer learning scenarios. We conduct extensive empirical evaluation on a multitude of benchmarks and show the effectiveness of our proposed baselines when compared with sophisticated state-of-the-art methods. The code is \href{https://github.com/OatmealLiu/MSc-iNCD}{open source}.
\end{abstract}

%%%%%%%%% MAIN PAPER
\section{Introduction}
\label{sec:introduction}
Clustering unlabelled samples in a dataset is a long standing problem in computer vision, where the goal is to group samples into their respective semantic categories. Given, there could be multiple valid criteria (\eg, shape, size or color) that could be used to cluster data, Deep Clustering (DC)~\cite{Xie2015UnsupervisedDE} can at times lead to clusters without desired semantics. A more efficient alternative was proposed in the work of Novel Class Discovery (NCD)~\cite{Han2019LearningTD}, where the goal is to discover and learn new semantic categories in an unlabelled dataset by transferring prior knowledge from labelled samples of related yet disjoint classes (see Fig.~\ref{fig:setting_comparison}a). In other words, NCD can be viewed as unsupervised clustering guided by known classes. Due to its practical usefulness, the field of NCD has seen a tremendous growth, with application areas ranging from object detection~\cite{fomenko2022learning} to 3D point clouds~\cite{riz2023novel}.

\begin{wrapfigure}{R}{0.49\textwidth}
\centering
% \lesspace
\vspace{-0.9cm}
    \includegraphics[width=\linewidth]{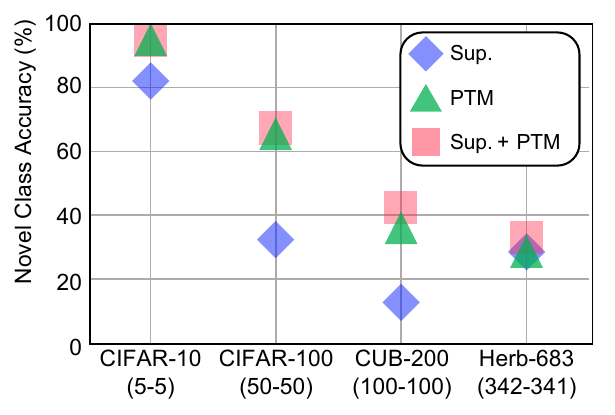}
    \vspace{-0.8cm}
    \caption{Comparison of traditional \textbf{Supervised} pre-training (Sup.) with self-supervised \textbf{pre-trained model} (PTM) initialization on the Novel Class Discovery.}
    \label{fig:ncd_pretrain_comp}
% \lesspace
\vspace{-0.7cm}
\end{wrapfigure}

A commonality in most of the \ncd methods~\cite{han2020automatically,fini2021unified} is that they rely on a reasonably large labelled dataset to learn good categorical and domain priors about the dataset. Thus, the success of these methods rely entirely on the availability of large labelled datasets, which might not always be guaranteed or can be expensive to collect. In this work we challenge the \textit{de facto} supervised pre-training step on a large labelled dataset for \ncd and show that supervised pre-training can be easily replaced by leveraging self-supervised pre-trained models (PTM), such as DINO~\cite{Caron2021EmergingPI}. PTMs being readily available off-the-shelf, it reduces the burden of pre-training on labelled data. As a part of a preliminary study, we compare supervised pre-training with PTMs and analyse their impact on the novel classes performance. As shown in Fig.~\ref{fig:ncd_pretrain_comp}, the PTMs achieve significantly better or at-par performance in comparison with the only supervised counterparts on all the datasets. Furthermore, when the PTMs are fine-tuned with supervised training on the labelled data, the performance is only marginally better. Note that the work in GCD~\cite{Vaze2022GeneralizedCD} used DINO as PTM, except it is used as initialization for the supervised training. Contrarily, we propose to entirely get rid of the supervised step.

Another striking drawback of the vast majority of \ncd methods, especially in ~\cite{han2020automatically,fini2021unified}, is that they assume access to the labelled dataset while discovering the novel (or \textit{new}) classes. Due to storage and privacy reasons the access to the labelled dataset can be revoked, which makes \ncd a very challenging problem. To address this, some very recent \cincdtitle (\cincd) methods~\cite{Roy2022ClassincrementalNC,Joseph2022NovelCD} have attempted to address \ncd from the lens of continual learning, by not accessing the labelled dataset when learning new classes (see Fig.~\ref{fig:setting_comparison}b). Albeit more practical than \ncd, the \cincd methods are still susceptible to catastrophic forgetting~\cite{French1999Catastrophic}, thereby impairing knowledge transfer from the labelled set to the unlabelled sets.

\begin{figure}[!t]
\begin{center}
\includegraphics[width=0.98\linewidth]{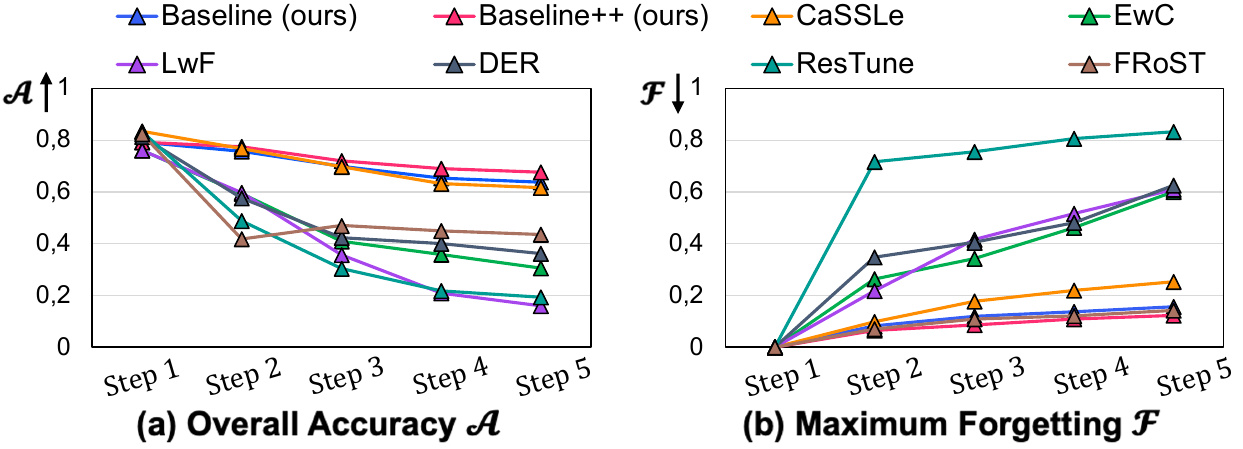}
\end{center}
% \lesspace
\vspace{-0.7cm}
\caption{Comparison of our proposed baselines with the incremental learning (EwC, LwF, DER), unsupervised incremental learning (CaSSLe), and \incd (ResTune, FRoST) methods on CIFAR-100. In each step 20 novel classes are learned. We report the Overall Accuracy and Maximum Forgetting.
}
\label{fig:sota_radar}
% \lesspace
\vspace{-0.7cm}
\end{figure}

In this work we aim to create a simple yet strong baseline for \cincd that can continually learn to cluster unlabelled data arriving in sessions, without losing its ability to cluster previously seen data. To this end, we propose \ours (see Fig.~\ref{fig:setting_comparison}c) that uses the DINO pre-trained ViT backbone, as a \textit{frozen} feature extractor, with a learnable linear \textit{cosine normalized} classifier~\cite{Hou2019LearningAU} on top. Every time an unlabelled set arrives, we simply train the task-specific classifier in a self-supervised manner, while keeping the backbone frozen. For testing we concatenate all the task-specific classifiers, yielding \textit{task-agnostic} inference. The simplicity of our approach lies in the decoupled training on task-specific data, while preserving performance across tasks. We characterize our \ours as \textit{frustratingly simple} as it \textit{neither} requires labelled data, \textit{nor} any specialized losses for preventing forgetting. Additionally, we propose \ourspp that stores discovered the novel class prototypes from the previous tasks to further reduce forgetting.

To verify the effectiveness of our proposed baselines, we compare with several state-of-the-art \cincd methods~\cite{liu2022residual,Roy2022ClassincrementalNC}, class-incremental learning methods (CIL)~\cite{kirkpatrick2017overcoming,li2017learning,buzzega2020dark} and unsupervised incremental learning (UIL)~\cite{Fini2021SelfSupervisedMA} methods adapted to the \cincd setting. In Fig.~\ref{fig:sota_radar} we plot the Overall Accuracy ($\mathcal{A}$) and Maximum Forgetting ($\mathcal{F}$) on CIFAR-100 for all the methods under consideration, where higher $\mathcal{A}$ and lower $\mathcal{F}$ is desired from an ideal method. Despite the simplicity, both the \ours and \ourspp surprisingly achieve the highest accuracy and least forgetting among all the competitors. Thus, our result sets a precedent to future \cincd methods and urge them to meticulously compare with our baselines, that are as simple as having a frozen backbone and a linear classifier.

In a nutshell, our \textbf{contributions} are three-fold: (\textbf{i}) We bring a paradigm shift in \ncd by proposing to use self-supervised pre-trained models as a new starting point, which can substitute the large annotated datasets. (\textbf{ii}) We, for the first time, highlight the paramount importance of having strong baselines in \cincd, by showcasing that simple baselines if properly implemented can outperform many state-of-the-art methods. To that end, we introduce two baselines (\ours and \ourspp) that are simple yet strong. (\textbf{iii}) We run extensive experiments on multiple benchmarks and for longer incremental settings.

To foster future research, we release a modular and easily expandable \href{https://github.com/OatmealLiu/MSc-iNCD}{PyTorch repository} for the \cincd task, that will allow practioners to replicate the results of this work, as well as build on top of our strong baselines.

\section{Related Work}
\label{sec:related_work}

\noindent
% \textbf{\ncdtitle (\ncd)}
\textbf{\ncdtitle (\ncd)} was formalized by \cite{Han2019LearningTD} with the aim of alleviating the innate ambiguity in deep clustering~\cite{Chang2017DeepAI,Dizaji2017DeepCV,Xie2015UnsupervisedDE,Yang2016TowardsKS,Yang2016JointUL} and enhancing the clustering ability of novel classes in an unlabelled dataset, by leveraging the prior knowledge derived from related labelled samples~\cite{Hsu2017LearningTC,Hsu2019MulticlassCW,Han2019LearningTD}. Many of the recent \ncd works utilize a joint training scheme that assumes access to both labelled and unlabelled data concurrently to exploit strong learning signal from the labelled classes~\cite{han2020automatically,Zhong2020OpenMixRK,fini2021unified,Jia2021JointRL,zhong2021neighborhood,Zhao2021NovelVC,Vaze2022GeneralizedCD,Fei2022XConLW,Yang2022DivideAC}. 

Keeping in mind the data regulatory practices, the \ncd community has been paying more attention to the problem of \incdtitle (\incd)~\cite{liu2022residual} where the access to the labelled (or base) dataset is absent during the discovery stage. Unlike \incd, FRoST~\cite{Roy2022ClassincrementalNC} and NCDwF~\cite{Joseph2022NovelCD} investigate a more realistic yet challenging setting known as \cincdtitle (\cincd), where task-id information is not available during inference. However, all the \cincd methods so far have investigated learning in short incremental scenarios (2 steps in \cite{Roy2022ClassincrementalNC} and 1 step in \cite{Joseph2022NovelCD}). Differently, we explore a more realistic setting of longer incremental setting (up to 5 steps) and show that many existing \cincd methods deteriorate in such settings.

Importantly, staying aligned with the original motivation of the \ncd and GCD paradigm -- \textit{discovering new classes by leveraging prior knowledge} -- we propose a new direction to tackle the \cincd problem, \ie, by solely leveraging the prior knowledge learned from self-supervised PTMs (\eg, DINO~\cite{Caron2021EmergingPI}), as opposed to relying on a large amount of expensive highly related \textit{labelled} data.
% or \textit{noisy web-scale} (\eg, CLIP~\cite{Radford2021LearningTV}

\noindent
\textbf{Class-incremental Learning (CIL)}~\cite{masana2022class} aims to train a model on a sequence of tasks with access to labelled data only from the current task, while the model's performance is assessed across all tasks it has encountered to date. Notably, the \il methods~\cite{kirkpatrick2017overcoming,Rebuffi2016iCaRLIC,li2017learning,buzzega2020dark} are devised with a dual objective of mitigating \textit{\forget}~\cite{French1999Catastrophic} of the model's knowledge on the previous tasks, while concurrently enabling it to learn new ones in a flexible manner. To overcome the need of labelled data, unsupervised incremental learning (UIL)~\cite{Fini2021SelfSupervisedMA,madaan2022representational,lin2022continual} have recently been proposed that aim to learn generalized feature representation via self-supervision to reduce forgetting. Different from UIL, that solely aims to learn a feature encoder, the \cincd methods additionally learn a classifier on top of the encoder to classify the unalabelled samples.

Moreover, as shown in the \cincd method FRoST~\cite{Roy2022ClassincrementalNC}, due to the differences in the learning objectives of \cincd during the supervised pre-training and unsupervised novel class discovery stages, learning continuously is more challenging than the supervised CIL setting. Our proposed baselines attempt to mitigate this issue with \compcnlong of the classifier weights, frozen backbone and feature replay using prototypes, thus greatly simplifying \cincd.

\section{Method}
\label{sec:method}
\noindent\textbf{Problem Formulation.} As illustrated in Fig.~\ref{fig:setting_comparison}c, a \cincd model is trained continuously over $T$ sequential \ncd tasks, each of which, $\mathcal{T}\tst$, presents an unlabelled data set $\data\tst = \{\vx\tst_n\}^{N\tst}_{n=1}$ with $N\tst$ instances containing $C\tst$ novel classes that correspond to a label set $\Y\tst$. As in prior works~\cite{troisemaine2023novel}, we assume that novel classes in $\data\tsi$ and $\data\tsj$ are disjoint, \ie, $\Y\tsi \cap \Y\tsj = \emptyset$. Following the \ncd literature, we assume the number of novel classes $C\tst$ at each step is known as \textit{a priori}.
\begin{wrapfigure}{R}{0.55\textwidth}
    \vspace{-0.7cm}
    \centering
    \includegraphics[width=\linewidth]{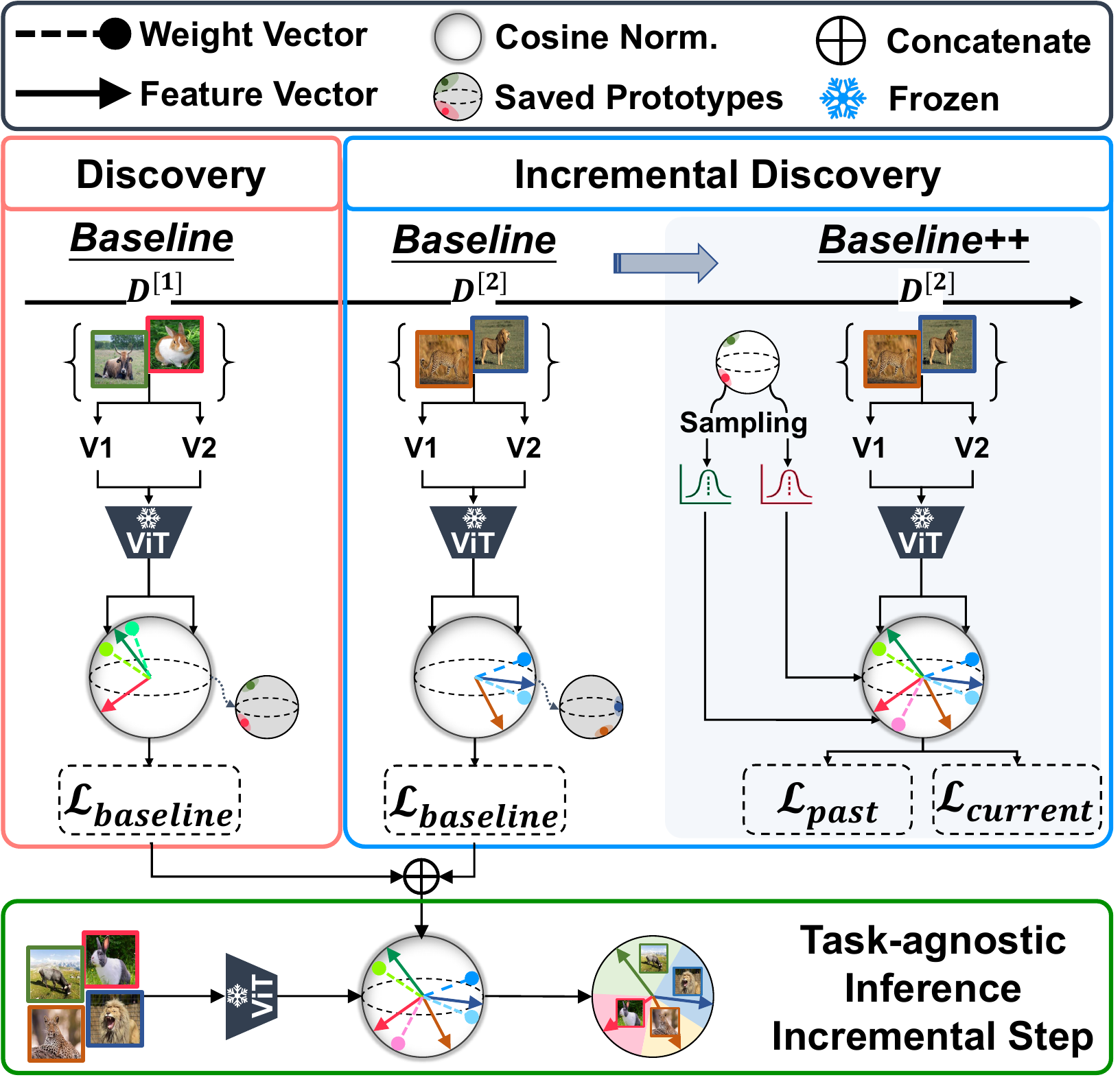}
    \lesspace
    \caption{
    Overview framework of the proposed methods \ours and \ourspp for class-iNCD task.
    }
    \label{fig:method_framework}
    % \vspace{-0.7cm}
    \lesspace
\end{wrapfigure}
During each discovery step $t$, we only have access to $\data\tst$. The aim of \cincd is to discover semantically meaningful categories in $\data\tst$ and accurately group the instances into the discovered clusters, without compromising its performance on the instances from $\data\tsone$ to $\data\tstminus$. In other words, a \cincd model comprises a unified mapping function $f \colon \X \to \bigcup_{t=1}^{T} \Y\tst$ that can group any test image $\vx$ into the categories $\bigcup_{t=1}^{T} \Y\tst$ discovered from the unlabelled task sequence $\taskb=\{\task\tsone, \task\tstwo, \cdots, \task\tsend\}$ without the help of task-id (\ie, task agnostic inference).

\subsection{Overall Framework}
\label{sec:method_overall_framework}
In this work our goal is to address \cincd by leveraging the priors learned by a self-supervised pre-trained model (PTM). To this end we propose a strong baseline called \ours that internally uses the PTM. As illustrated in Fig.~\ref{fig:method_framework}, the proposed \ours is marked by two steps -- (\textbf{i}) an initial \textbf{discovery step} (see pink box), where task-specific classifier is learned to discover the novel classes contained in $\data\tsone$ for the first task $\task\tsone$ with a clustering objective ($\mathcal{L}_\mathrm{baseline}$). Pseudo per-class prototypes are computed and stored; and (\textbf{ii}) it is followed by an \textbf{incremental discovery step} (see blue box), where \ours conducts the same discovery training, after which \textbf{task-agnostic inference} (see green box) is performed by simply concatenating the two learned task-specific classifiers. \ourspp further fine-tunes the concatenated classifier with $\mathcal{L}_\mathrm{past}$ and $\mathcal{L}_\mathrm{current}$ using the stored class prototypes to strength class-discrimination among tasks.
In the following sections, we first present a comprehensive overview of \ours. Additionally, we introduce an advanced variant of \ours, named \ourspp, which incorporates feature replay to further mitigate the issue of forgetting.

\noindent
\textbf{Discovery Step.}
In the introductory discovery task $\task\tsone$ (see Fig.~\ref{fig:method_framework}), we learn a mapping function $f\tsone \colon \X\tsone \to \Y\tsone$ in a self-supervised manner (\ie, using the Sinkhorn-Knopp cross-view pseudo-labelling~\cite{Caron2020UnsupervisedLO}) to discover the $C\tsone$ categories contained in the given unlabelled data set $\data\tsone$. The mapping function $f\tsone = h\tsone \circ g$ is modeled by a \textit{frozen} feature extractor $g(\cdot)$ and a \textit{Cosine Normalized} linear layer $h\tsone(\cdot)$ as task-specific classifier. The $g(\cdot)$ is initialized by the PTM weights $\theta_{g}$~\cite{Caron2021EmergingPI}, while $h\tsone(\cdot)$ is randomly initialized. In other words, only the classifier $h\tsone(\cdot)$ weights are learned during this step.

\noindent
\textbf{Incremental Discovery Step.}
After the first discovery step, $\data\tsone$ is discarded, and access to only $\data\tstwo$ is given in the first \textit{incremental} discovery step $\task\tstwo$ (see Fig.~\ref{fig:method_framework}). Same as the first step, we train a task-specific mapping function modeled by $f\tstwo = h\tstwo \circ g$. The $h\tstwo$ is newly initialized for the $C\tstwo$ novel classes of $\data\tstwo$, while the \textit{frozen} $g$ is shared across tasks. Thanks to the \textit{frozen} feature extractor and \textit{\compcntitle} (\compcn), \ours easily forms a unified model $f\tstotwo = h\tstotwo \circ g$ by sharing the feature extractor $g$, and concatenating the two task-specific heads $h\tstotwo(\cdot) = h\tsone(\cdot) \oplus h\tstwo(\cdot)$ for task-agnostic inference.

\noindent
\textbf{Task-agnostic Inference.} After training for $T$ steps, the inference on the test samples, belonging to any class presented in $\taskb$, is carried out with the final unified model $f\tstoend = h\tstoend \circ g$ in a task-agnostic manner (see Fig.~\ref{fig:method_framework}).

\subsection{Why Use Self-supervised Pre-trained Models?}
\begin{wraptable}{r}{0.49\textwidth}
    % \tablestyle{1.0pt}{0.1}
    \tiny
    \centering
    \vspace{-0.8cm}
    \caption{Analysis of \ncd accuracy using the same backbone (\vitbsixteen) with different pre-training settings.}
    \label{tab:ncd_pretraining}
    \begin{tabular}{lcccc}
        \toprule
        \multirow{2}{*}{Pre-training} & CIFAR-10 & CIFAR-100 & CUB-200 & \multirow{2}{*}{Avg. ($\Delta$)}\\
        & (5-5) & (50-50) & (100-100)&\\
        % \cline{2-5}
        \hline
        Supervised & 82.1 & 32.4 & 12.8 & 42.4\\
        PTM-DINO & \textbf{95.0} & 65.6 & 36.1 & 65.6 (\textcolor{darkgreen}{+23.2\%})\\
        PTM-DINO + Supervised & 94.5 & \textbf{67.2} & \textbf{42.5} & \textbf{68.1} (\textcolor{darkgreen}{+25.7\%})\\
     \bottomrule
    \end{tabular}
    % \vspace{-0.8cm}
    \lesspace
\end{wraptable}

Before delving into the specifics of our method, we first validate the benefits of leveraging self-supervised PTMs for \ncd, where supervised pre-training is the standard practice. Specifically, we conduct experiments with our \ours method under traditional \ncd setting and splits~\cite{fini2021unified} on three benchmarks (CIFAR-10, CIFAR-100 and CUB-200), comparing three pre-training strategies: (i) supervised pre-training on the labelled set starting from a randomly initialized model (as in Fig.~\ref{fig:setting_comparison}a), (ii) self-supervised PTM initialization (\eg, DINO~\cite{Caron2021EmergingPI}, a \textit{self-supervised} model) (as in Fig.~\ref{fig:setting_comparison}c), and (iii) supervised fine-tuning starting from PTM initialization. After this step the novel classes are discovered in the unlabelled set. In Tab.~\ref{tab:ncd_pretraining} we can see that the PTM-DINO, a model trained without any supervision, performs significantly better in discovering novel classes compared to the supervised counterpart (by +23.2\%), which is trained on the highly related base classes. This demonstrates that the original motivation of using a highly related labelled set to aid NCD~\cite{Han2019LearningTD} is clearly suboptimal when compared with self-supervised pre-training on a rather larger dataset. Additionally, fine-tuning PTM-DINO on the labelled samples only gives limited accuracy gain, with the PTM-DINO performing reasonably at-par (-2.5\%). Guided by these observations, we propose using strong PTMs (\eg, DINO~\cite{Caron2021EmergingPI}) with Vision Transformers (ViT)~\cite{Dosovitskiy2020AnII} as a new starting point for \ncd and \cincd, thereby eliminating the dependence on the labelled data.

\subsection{Strong Baselines for \cincd}
\label{sec:method_discovery}
In this section we detail the proposed methods, \ours and \ourspp, for solving the \cincd task. Both the baselines use PTMs, as backbone, that are general purpose and publicly available. Additionally, the \ourspp uses latent feature replay. The baselines have been designed to preserve stability on the \textit{past} novel classes, while being flexible enough to discover the novel classes in the \textit{current} task.

% \vspace{-.05in}
\begin{tcolorbox}[colback=gray!30,halign=center,valign=center,height=0.2in]
\textbf{Baseline}
\end{tcolorbox}
% \vspace{-.05in}

% \subsubsection{Baseline}
\noindent
\textbf{Self-supervised Training for Discovery.} Starting from a frozen feature extractor $g$, initialized with the weights from DINO~\cite{Caron2021EmergingPI}, we optimize a \textit{self-supervised} clustering objective to directly discover the novel categories at each step. In details, first we randomly initialize a learnable linear layer $h\tst$ as the task-specific classifier for the $C\tst$ novel classes contained in the unlabelled set $\data\tst$. To learn the task-specific network $f\tst$ for discovery, \ours employs the Sinkhorn-Knopp cross-view pseudo-labeling algorithm~\cite{Caron2020UnsupervisedLO}. We optimize a \textit{swapped} prediction problem, where the `code' $\vy_1$ of one view is predicted from the representation of another view $\vz_2$, derived from the same image $\vx$ through different image transformations, and vice-versa:

\begin{equation}
% \small
    \mathcal{L}_{\ours}= \ell(\vz_2, \vy_1) + \ell(\vz_1, \vy_2)
\label{eqn:ce}
\end{equation}
where $\ell(\cdot, \cdot)$ is the standard cross-entropy loss. We obtain the codes (or \textit{soft-targets}) $\vy_1$ and $\vy_2$ by using the \sinkhorn algorithm. 
%Refer to the supplement for details. 
Note that, we freeze the entire feature extractor $g$ during optimizing $\mathcal{L}_\ours$ as a straightforward way to prevent catastrophic forgetting.

\noindent
\textbf{Multi-step Class-Incremental Discovery.} Our ultimate goal is to learn a unified mapping function $f\tstoend \colon \X \to \bigcup_{t=1}^{T} \Y\tst$. If all the training data are available, an ideal clustering objective for $f\tstoend$ can be achieved by minimizing an adequate loss $\mathcal{L}\tstoend$ at the end of the task sequence:
\begin{equation}
% \small
% \vspace{-2mm}
\label{eqn:ideal}
    \mathcal{L}\tstoend =
    \E_{\task\tst\sim\taskb}
    % \bigl[
    %\E_{(\vx, \hat{\vy})\sim\task\tst} \ell(\hat{\vy}, f\tstoend(\vx)),
    % \bigr] 
    \mathcal{L}\tst
% \vspace{-2mm}
\end{equation}
However, due to the data unavailability of past tasks in \cincd, we can only pursue an approximation of this ideal joint objective defined by Eq.~\ref{eqn:ideal}. In this work, unlike most of the CIL solutions~\cite{wang2023comprehensive}, we pursue a better approximation from a new perspective: balancing the individual clustering objectives in each task to a unified importance. To be more specific, the proposed \ours adopts \textit{frozen feature extractor} with \textit{cosine normalized classifier} to unify the clustering objectives across tasks.
%, while \ourspp further leverages \textit{\compfrlong} training to achieve a better balance for all tasks. We introduce each of the proposed component below.
% The \textit{past-current} trade-off in this approximation incurs the notorious \textit{\forget} problem in \il tasks~\cite{wang2023comprehensive}.

\noindent\textit{\textbf{Frozen Feature Extractor.}} In \ours we freeze the entire PTM $g$ by introducing $\Vert \theta_{g}\tst - \theta_{g}\tstminus \Vert^{2} = 0$, $ t \in \{1, \ldots, T\}$ as a constraint. This enables us to leverage the power of the generalist PTM $g$ for all tasks \textit{equally}, without  introducing bias towards any particular task, \ie, avoiding the \textit{model drift} issue in CIL literature~\cite{wu2022class}.

\noindent\textit{\textbf{Cosine Normalization.}} The frozen feature extractor not only preserves the powerful prior knowledge from the pre-training data, but also maintains the cooperative mechanism between $g$ and each individual classifier $h\tst$. With the stable cooperative mechanism, the test data can be directly routed to the corresponding task-specific function network $f=h\tst \circ g$, if the task-id $t$ is available. However, the task-id is not available in \cincd. To achieve simple task-agnostic inference, we propose to apply \textit{\compcntitle} (\compcn)~\cite{Luo2017CosineNU,Hou2019LearningAU} on each individual linear classifier $h\tst$. This enables the learned classifiers to output scores of the same scale, avoiding imbalance between the past and current novel classes. 

Formally, given an input vector $\vx$, the L2 normalization operation can be defined as $\widetilde{\vx} = L2Norm(\vx) = \nicefrac{\vx}{\Vert \vx \Vert} = \nicefrac{\vx}{\sqrt{\vx\vx^{T} + \epsilon}}$, where $\epsilon$ is a small value to avoid division by zero and is set to 1e$^{-12}$ in this work.
%$\Vert \vx \Vert=\sqrt{\vx\vx^{T} + \epsilon}$ is L2-norm calculation ($\epsilon$ is a small value to avoid division by zero, which is set to 1e-12 in all experiments of this work). 
At every discovery step, $L2Norm(\cdot)$ is continuously applied to both the input feature embedding $\vz$ and each weight vector $\theta_{h}^{i}$ of the task-specific linear classifier $h\tst$. $\theta_{h}^{i} \in \R^{k}$ is the $i$-th column of the classifier weight matrix $\theta_{h}$, corresponding to one semantic cluster. Consequently, the $i$-th output logit from the classifier is then computed as:
\begin{equation}
% \small
\label{eqn:normlogits}
    \vl^{i} = \widetilde{\theta}_{h}^{i T} \widetilde{\vz} = \frac{\theta_{h}^{i T} \vz}{\Vert \theta_{h}^{i} \Vert \Vert \vz \Vert} = cos(\theta_{h}^{i})
\end{equation}
where $\Vert \theta_{h\tst}^{i} \Vert = \Vert \vz \Vert = 1$ and $cos(\theta_{h\tst}^{i})$ is the cosine similarity between the feature vector $\vz$ and the $i$-th weight vector $\theta_{h\tst}^{i}$. We thus use the term \textit{\compcn} for this operation.
%Therefore, the output calculation of $h\tst$ can be interpreted as the cosine similarity comparison between the input feature embedding and its weight vectors, hence the term \textit{\compcn} for this operation. 
The magnitude of the output logits $\vl$ is thereby unified to the same scale $[-1,1]$ for all classifiers from different steps.

\noindent
\textbf{Task-agnostic Inference.} Having the balanced classifier weights, we can then build a unified classification head $h\tstoend$ by simply concatenating the task-specific heads learned at each step $h\tstoend = h\tsone \oplus h\tstwo \oplus \ldots \oplus h\tsend$. By means of the frozen feature extractor and feature normalization, all the feature embedding $\widetilde{\vz}$ are mapped to the uniform feature space under the same scale. Incorporating with the normalized classifier weights in $h\tstoend$, task-agnostic inference can be fairly achieved using $f\tstoend=h\tstoend \circ g$ for all the discovered classes so far.

% \subsubsection{Baseline++}
% \vspace{-.05in}
\begin{tcolorbox}[colback=gray!30,halign=center,valign=center,height=0.2in]
\textbf{Baseline++}
\end{tcolorbox}
% \vspace{-.05in}

To take full advantage of the stable feature extractor, we propose \ourspp that additionally uses the learned model $f\tstminus = h\tstminus \circ g$ to compute the pseudo per-class feature prototypes $\protomean_{\hat{c}\tstminus}$ and variances $\protovar_{\hat{c}\tstminus}$ as \textit{proxies} for the novel classes discovered from the previous task $\mathcal{T}\tstminus$. For the subsequent tasks, features drawn from the Gaussian distribution, constructed with the stored $\protomean_{\hat{c}\tstminus}$ and $\protovar_{\hat{c}\tstminus}$, are replayed to reduce forgetting in the classifiers. We call this simplified replay mechanism as \textit{\compfrtitle} (\compfr) (see Fig.~\ref{fig:method_framework}), which we describe next.

\noindent
\textbf{Knowledge Transfer with Robust Feature Replay (\compfr).} 
At each previous discovery step $t \in \{1, \ldots, T-1\}$, \ourspp computes and stores a set $\Mu\tst = \{\mathcal{N}(\protomean_{\hat{c}_{j}\tst}, \protovar_{\hat{c}_{j}\tst})\}_{j=1}^{\classes\tst}$ that contains pseudo per-class feature prototype distributions derived from the unlabelled set $\data\tst$. Here, $\protomean_{\hat{c}_{j}\tst}$ and $\protovar_{\hat{c}_{j}\tst}$ are the calculated mean and variance of the feature embedding predicted by the task-specific model $f\tst$ as pseudo class $\hat{c}_{j}\tst$. Since the feature prototype set $\Mu\tst$ can represent and simulate the novel classes discovered at each previous step, \ourspp can further train the concatenated model $f\tstoend=h\tstoend \circ g$ by replaying the per-class features sampled from the saved Gaussian distributions in $\{\Mu\tsone, \ldots, \Mu\tstminuscap\}$ with the objective defined as:
\begin{equation}
% \small
\label{eqn:loss_past}
    % \begin{split}
    \mathcal{L}_\mathrm{past}=- \E_{\Mu\tst \sim \Mu\tstoendminusone}\E_{(\vz^{\hat{c}\tst}, \hat{\vy}^{\hat{c}\tst}) \sim \mathcal{N}(\protomean_{c\tst}, \protovar_{c\tst})}
    % &
    \sum_{j=1}^{\classes\tst}\hat{\vy}^{\hat{c}_{j}\tst} \log \sigma(\frac{h\tstoend(\vz^{\hat{c}_{j}\tst})}{\tau})
    % \end{split}
\end{equation}

where, $\sigma (\cdot)$ is a softmax function and $\tau$ is the temperature. By optimizing the objective defined in Eq.~\ref{eqn:loss_past}, \ourspp can better approximate the ideal objective defined in Eq.~\ref{eqn:ideal} by simulating the past data distribution. Furthermore, to maintain the clustering performance for the current novel classes in $\data\tsend$, we also transfer the knowledge from the current task-specific head $h\tsend$ to $h\tstoend$. In details, using the pseudo-labels $\hat{\vy}_{i}\tsend$ computed by the learned $f\tsend$, we can build a pseudo-labelled data set $\data_{PL}\tsend = \{\vx_{i}\tsend, \hat{\vy}_{i}\tsend\}_{i=1}^{N\tsend}$. The task-specific knowledge stored in the pseudo-labels can be then transferred to the unified classifier by optimizing the following objective:
\begin{equation}
% \small
\label{eqn:loss_now}
% \begin{split}
\mathcal{L}_\mathrm{current}=-\E_{(\vx\tsend, \hat{\vy}\tsend) \sim \data_{PL}\tsend}
    \sum^{C\tsend}_{j=1} \hat{y}_{j}^{c\tsend} \log \sigma(\frac{h\tstoend(g(\vx^{c\tsend}))}{\tau}).
% \end{split}
\end{equation}
The final \textit{past-current} objective for \compfr training at step $T$ of \ourspp is formulated as:
\begin{equation}
% \small
\label{eqn:loss_all}
    \mathcal{L}_\mathrm{\ourspp} = \mathcal{L}_\mathrm{past} + \mathcal{L}_\mathrm{current}
\end{equation}

\section{Experiments}
\label{sec:experiments}
\subsection{Experimental Settings}
\label{sec:expt_setup}
\noindent
\textbf{Datasets and Splits.} We conduct experiments on three generic image recognition datasets and two fine-grained recognition datasets: CIFAR-10 (C10)~\cite{krizhevsky2009learning}, CIFAR-100 (C100)~\cite{krizhevsky2009learning}, TinyImageNet-200 (T200)~\cite{le2015tiny}, CUB-200 (B200)~\cite{wah2011caltech} and Herbarium-683 (H683)~\cite{Tan2019TheHC}. Although the PTM (DINO) used in our baselines and the methods we compared was pre-trained without labels,there's a potential for category overlap between the pre-training dataset (ImageNet~\cite{Deng2009ImageNetAL}) and C10, C100, and T200. To ensure a equitable evaluation, we include B200 and H683 datasets. Notably, B200 shares only two categories with DINO's pre-training dataset (ImageNet), whereas H683 has no overlap whatsoever. For each dataset, we adopt two strategies (two-step and five-step) to generate the task sequences, where the total classes and corresponding instances of training data are divided averagely for each step. The test data are used for evaluation. Detailed data splits are provided in the supplementary material.

\noindent
\textbf{Evaluation Protocol.} We evaluate all the methods in \cincd using the \textbf{task-agnostic} evaluation protocol~\cite{Roy2022ClassincrementalNC}. Specifically,  we do not know the task ID of the test sample during inference, and the network must route the sample to the correct segment of the unified classifier.

\noindent
\textbf{Evaluation Metrics.}
We report two metrics: maximum forgetting $\forgetting$ and overall discovery accuracy (or clustering accuracy~\cite{Roy2022ClassincrementalNC}) $\accuracy$ for all discovered classes by the end of the task sequence. $\forgetting$ measures the difference in clustering accuracy between the task-specific model $f\tsone$ and the unified model $f\tstoend$ (at the last step) for samples belonging to novel classes discovered at the first step. $\accuracy$ is the clustering accuracy from the unified model $f\tstoend$ on instances from all the novel classes discovered by the end of the sequence.

\subsection{Implementation Details}
\label{sec:expt_implementation}

\noindent
\textbf{\ours and \ourspp.}
By default, \vitbsixteen~\cite{Dosovitskiy2020AnII} is used as the backbone $g$ with DINO~\cite{Caron2021EmergingPI} initialization for all data sets. The 768-dimensional output vector $\vz \in \R^{768}$, from the $[CLS]$ token is used as the deep features extracted from a given image. $g$ is frozen during training. Following the backbone, one \textit{\cosnormed} linear layer (without bias) is randomly initialized as the task-specific classifier $h\tst$ with $\classes\tst$ output neurons. Soft pseudo-labels self-supervised are generated using the \sinkhorn~\cite{Cuturi2013SinkhornDL,Caron2020UnsupervisedLO} algorithm with default hyper-parameters (number of iterations = 3 and $\epsilon=0.05$).

\noindent
\textbf{Training.}
At each step, we train the model for 200 epochs on the given unlabelled data set $\data\tst$ with the same data augmentation strategy~\cite{Chen2020ASF} in all the experiments. After the discovery stage, \ourspp further conducts \compfr training on the unified model $f\tstot$ for 200 epochs. A cosine annealing learning rate scheduler with a base rate of 0.1 is used. The model is trained on mini-batches of size 256 using SGD optimizer with a momentum of 0.9 and weight decay $10^{-4}$. The temperature $\tau$ is set to 0.1.

\subsection{Analysis and Ablation Study}
\label{sec:expt_self_analysis}
%%%%%%%%%%%%%%%% here we compare our methods with the constructed lower and upper bounrds.
\noindent
\textbf{Comparison with Reference Methods.} We first establish reference methods using K-means~\cite{Arthur2007kmeansTA} and joint training scheme ($\upperb$, based on \ours but access to the previous training data is given)~\cite{li2017learning}, respectively. 
\begin{wraptable}{r}{0.52\textwidth}
    \tablestyle{1.0pt}{0.9}
    \small
    \vspace{-0.7cm}
    \caption{Comparison of our proposed baselines with reference methods on two task splits of C10, C100 and T200.}
    \label{tab:expt_upperlower}
    \lesspace
    \begin{center}
        \begin{tabular}{cl|cc|cc|cc}
            \toprule
            
            & Datasets & \multicolumn{2}{c|}{C10} & \multicolumn{2}{c|}{C100} & \multicolumn{2}{c}{T200} \\
            
            & Methods & $\forgetting\downarrow$ & $\accuracy\uparrow$ & $\forgetting\downarrow$ & $\accuracy\uparrow$ & $\forgetting\downarrow$ & $\accuracy\uparrow$ \\
            \hline
    
            \multirow{5}{*}{\rotatebox[origin=c]{90}{Two-step}}&Kmeans~\cite{Jain2008DataC5} & 93.9 & 87.3 & 68.2 & 56.7 & 62.0 & 47.1 \\
            &Joint (frozen) & 4.9 & 92.1 & 5.3 & 61.8 & 3.3 & 51.1 \\
            &Joint (unfrozen) & \textbf{0.8} & \textbf{92.4} & \textbf{2.5} & \textbf{65.2} & 2.3 & \textbf{56.5} \\
            
            \cline{2-8}
            
            &\ours & 8.5 & {89.2} & 6.7 & {60.3} & 4.0 & {54.6} \\
            &\ourspp & 4.5 & {90.9} & 6.6 & {61.4} & \textbf{0.2} & {55.1} \\
            
            \toprule
            \multirow{5}{*}{\rotatebox[origin=c]{90}{Five-step}}&Kmeans~\cite{Jain2008DataC5} & 99.1 & 82.1 & 76.3 & 54.3 & 66.0 & 52.9 \\
            &Joint (frozen) & 5.1 & 93.8 & 10.5 & 68.6 & 1.8 & 57.8 \\
            &Joint (unfrozen) & \textbf{1.5} & \textbf{97.5} & \textbf{5.9} & \textbf{74.9} & 3.0 & \textbf{60.7} \\ 
            \cline{2-8}
            &\ours & 8.2 & {85.4} & 15.6 & {63.7} & 9.2 & {53.3} \\
            &\ourspp & 7.6 & {91.7} & 12.3 & {67.7} & \textbf{1.6} & {56.5} \\
            
            \bottomrule
        \end{tabular}
    \end{center}
    \lesspace
% \end{table}
\end{wraptable}
To further enhance the upper reference performance, we unfreeze the last transformer block during training on joint data sets, which is referred as to $\upperbpp$ method. As shown in Tab.~\ref{tab:expt_upperlower}, the joint training methods slightly outperform our baselines on all data sets and splits, as they can jointly optimize the ideal objective defined in Eq.~\ref{eqn:ideal} using the given access to all training data. Nonetheless, our baselines perform nearly as well as the joint training methods, indicating limited benefits from access to all unlabelled data under \cincd and the effectiveness of our baselines.

\begin{table}[!t]
% \begin{wraptable}{r}{0.52\textwidth}
    % \tablestyle{1.0pt}{0.7}
    % \setlength{\tabcolsep}{3.2pt}
    % \vspace{-0.7cm}
    \caption{Self-ablation analysis of the proposed components on two task splits of C10, C100 and T200.}
    \label{tab:expt_selfablation}
    \vspace{-0.3cm}
    \begin{center}
        \begin{tabular}{cccccc|cc|cc}
            \toprule
            
            &&\multicolumn{2}{l}{Datasets} & \multicolumn{2}{c|}{C10} & \multicolumn{2}{c|}{C100} & \multicolumn{2}{c}{T200} \\
            
            &&\compcn & \compfr  & $\forgetting\downarrow$ & $\accuracy\uparrow$ & $\forgetting\downarrow$ & $\accuracy\uparrow$ & $\forgetting\downarrow$ & $\accuracy\uparrow$ \\
            \hline
            \multirow{4}{*}{\rotatebox[origin=c]{90}{Two-step}}&(a)&\Checkmark & \Checkmark & \textbf{4.5} & \textbf{90.9} & {6.6} & \textbf{61.4} & \textbf{0.2} & \textbf{55.1} \\
            &(b)&\Checkmark & \XSolidBrush & 8.5 & 89.2 & 6.7 & 60.3 & 4.0 & 54.6 \\
            &(c)&\XSolidBrush & \Checkmark & 8.2 & 80.2 & \textbf{5.1} & 54.1 & 3.3 & 38.9\\
            &(d)&\XSolidBrush & \XSolidBrush & 16.1 & 74.3 & 7.3 & 50.1 & 4.3 & 33.2 \\
            
            \hline
    
            \hline
            \multirow{4}{*}{\rotatebox[origin=c]{90}{Five-step}}&(a)&\Checkmark & \Checkmark & {7.6} & \textbf{91.7} & \textbf{12.3} & \textbf{67.7} & {1.6} & \textbf{56.5} \\
            &(b)&\Checkmark & \XSolidBrush & 8.2 & 85.4 & 15.6 & 63.7 & 9.2 & 53.3 \\
            &(c)&\XSolidBrush & \Checkmark &\textbf{6.3} & 90.7 & 14.3 & 58.2 & \textbf{0.7} &49.7 \\
            &(d)&\XSolidBrush & \XSolidBrush & 10.9 & 80.3 & 16.6 & 49.1 & 8.1 & 41.9 \\
            \bottomrule
        \end{tabular}
    \end{center}
    \vspace{-0.9cm}
\end{table}

\noindent
\textbf{Ablation on Proposed Components.} We further present an ablation study on the individual core components of our baseliens, namely \compcn and \compfr. Results are shown in Tab.~\ref{tab:expt_selfablation}. It is noticeable from the results that \compcn plays a substantial role in enhancing the overall accuracy of our proposed baselines (refer to \ours: \textit{(b) v.s. (d)} and \ourspp: \textit{(a) v.s. (c)}). This is attributed to its unification capability to effectively address the issue of that the weight vectors with significant magnitudes in $f\tstoend=h\tstoend \circ g$ always dominating the prediction. On the other hand, \compfr can improve the overall accuracy and mitigate the forgetting at the end of each task sequence (refer to \textit{(a) v.s. (b)} and \textit{(c) v.s. (d)}). Of particular note is that the performance gain attained by using \compfr is more significant when dealing with longer task sequences (refer to the \textit{upper half v.s. lower half} in Tab.~\ref{tab:expt_selfablation}).
\ourspp (a) equipped with both \compcn and \compfr achieves the best overall accuracy and the least forgetting.

%%%%%%%%%%%%%%%% here we analyze the effectiveness of different large-scale pre-trained models.
% \begin{table}[!t]
\begin{wraptable}{r}{0.48\textwidth}
    \tablestyle{1.pt}{0.9}
    \vspace{-0.45cm}
    \small
    \lesspace
    \caption{Ablation of architectures and pre-training strategies of PTMs on five-step splits of C10, C100 and T200.}
    \label{tab:expt_backboneablation}
    \vspace{-0.5cm}
    \begin{center}
        \begin{tabular}{lcccccc}
            \toprule
            % Five-step Split ($\searrow$) &\multicolumn{6}{c}{\ours}\\
            &\multicolumn{6}{c}{\textbf{\ours}}\\
            Datasets & \multicolumn{2}{c}{C10} & \multicolumn{2}{c}{C100} & \multicolumn{2}{c}{T200}\\
    
            % & \multicolumn{2}{c}{$\task\tstofive$} & \multicolumn{2}{c}{$\task\tstofive$} & \multicolumn{2}{c}{$\task\tstofive$} \\
            
            Backbones& $\forgetting \downarrow$ & $\accuracy \uparrow$ & $\forgetting \downarrow$ & $\accuracy \uparrow$ & $\forgetting \downarrow$ & $\accuracy \uparrow$ \\
            \hline
            ResNet50-DINO & 37.5 & 45.8 & 16.4 & 38.5 & 10.1 & 24.7 \\
            ViT-B/16-DINO & 8.2 & 85.4 & \textbf{15.6} & \textbf{63.7} & \textbf{9.2} & \textbf{53.3}\\
            ViT-B/16-CLIP & \textbf{5.3}& \textbf{87.5} & 17.1 & 62.4 & 15.7 & 42.5 \\  
            
            \bottomrule
        \end{tabular} 
    \end{center}
    \vspace{-0.5cm}
\end{wraptable}
% \end{table}

\noindent
\textbf{Analysis of Pre-Trained Models (PTM)}. In Tab.\ref{tab:expt_backboneablation}, we present a comparison between different PTMs such as ResNet50~\cite{He2015DeepRL} and \vitbsixteen~\cite{Dosovitskiy2020AnII}, along with various pre-training strategies (CLIP~\cite{Radford2021LearningTV} and DINO~\cite{Caron2021EmergingPI}). Transformer architecture achieves superior performance owing to its discrimination ability~\cite{Naseer2021IntriguingPO}. CLIP pre-training achieves similar outcomes to DINO, demonstrating the effectiveness of strong PTM with a different pre-training strategy on web data.

\subsection{Comparison with the State-of-the-art Methods}
\label{sec:expt_sota_comparison}
For a comprehensive comparison, we adapt methods from closely related fields for state-of-the-art comparison. We adjust ResTune~\cite{liu2022residual} and FRoST~\cite{Roy2022ClassincrementalNC} to the multi-step \cincd setting from the closely related \incd field. We adapt three representative CIL methods: EwC~\cite{kirkpatrick2017overcoming}, LwF~\cite{li2017learning}, and DER~\cite{buzzega2020dark} to this self-supervised setting. Similarly, we adapt the \uil method, CaSSLe~\cite{Fini2021SelfSupervisedMA}, for incremental discovery. All adapted methods employ \vitbsixteen with the same DINO-initialization as a feature extractor. For the adapted CIL and \uil methods, the same self-training strategy is used as in our \ours method to prevent forgetting. All adapted methods unfreeze only the last transformer block of the feature extractor~\cite{Wu2019LargeSI,Boschini2022TransferWF}, except ResTune that unfreezes the last two blocks for model growing. More implementation details can be found in the supplementary material.

\begin{table}[!t]
    % \tablestyle{1.0pt}{0.9}
    \small
    % \vspace{-0.7cm}
    \caption{Comparison with the adapted state-of-the-art methods on two task splits of C10, C100, T200, B200, and H683 under \cincd setting with the same DINO-\vitbsixteen backbone. Overall accuracy and maximum forgetting are reported.}
    \label{tab:expt_sota}
    \vspace{-0.3cm}
    \begin{center}
        \begin{tabular}{cl|cc|cc|cc|cc|cc}
            \toprule
            
            & Datasets 
            & \multicolumn{2}{c|}{C10} 
            & \multicolumn{2}{c|}{C100} 
            & \multicolumn{2}{c|}{T200}
            & \multicolumn{2}{c|}{B200}
            & \multicolumn{2}{c}{H683} \\

            & Methods 
            & $\forgetting\downarrow$ 
            & $\accuracy\uparrow$ 
            & $\forgetting\downarrow$ 
            & $\accuracy\uparrow$ 
            & $\forgetting\downarrow$ 
            & $\accuracy\uparrow$ 
            & $\forgetting\downarrow$ 
            & $\accuracy\uparrow$ 
            & $\forgetting\downarrow$ 
            & $\accuracy\uparrow$ \\
            \hline
    
            \multirow{8}{*}{\rotatebox[origin=c]{90}{\textbf{Two-step}}}
            &EwC~\cite{kirkpatrick2017overcoming} & 32.4 & 79.0 & 42.5 & 43.9 & 27.2 & 33.3 & 18.1 & 25.5 & 13.8 & 25.1 \\
            &LwF~\cite{li2017learning} & 30.4 & 34.4 & 44.1 & 42.4 & 40.0 & 27.2 & 20.2 & 23.9 & 16.3 & 24.9\\
            &DER~\cite{buzzega2020dark} & 49.0 & 69.9 & 29.8 & 30.3 & 39.0 & 28.9 & 5.0 & 20.4 & 14.0 & 24.7\\
            &ResTune~\cite{liu2022residual} & 97.6 & 47.2 & 32.7 & 17.1 & 32.3 & 17.2 & 12.0 & 13.0 & 27.4 & 17.1\\
            &FRoST~\cite{Roy2022ClassincrementalNC} & \textbf{2.5} & 46.6 & \textbf{4.7} & 34.2 & 4.3 & 26.1 & \textbf{3.9} & 17.6 & 16.2 & 18.4 \\
            &CaSSLe~\cite{Fini2021SelfSupervisedMA} & 9.1 & 87.3 & 10.3 & 53.7 & 6.9 & 36.5 & 4.8 & 26.8 & 10.9 & 25.3 \\
            
            \cline{2-12}
            
            &\ours & 8.5 & \textbf{89.2} & 6.7 & \textbf{60.3} & 4.0 & \textbf{54.6} & 4.1 & \textbf{28.7} & 7.9 & \textbf{25.7} \\
            &\ourspp & 4.5 & \textbf{90.9} & 6.6 & \textbf{61.4} & \textbf{0.2} & \textbf{55.1} & 4.2 & \textbf{36.9} & \textbf{6.0} & \textbf{27.5}\\
            
            \toprule
            \multirow{8}{*}{\rotatebox[origin=c]{90}{\textbf{Five-step}}}
            &EwC~\cite{kirkpatrick2017overcoming} & 21.1 & 81.1 & 60.1 & 30.6 & 48.0 & 23.2 & 21.2 & 19.1 & 15.7 & 22.4\\
            &LwF~\cite{li2017learning} & 20.1 & 25.8 & 60.9 & 16.1 & 53.7 & 15.6 & 21.7 & 15.7 & 16.5 & 23.4\\
            &DER~\cite{buzzega2020dark} & 30.1 & 76.2 & 62.6 & 36.2 & 52.1 & 21.7 & 16.2 & 16.3 & 18.0 & 22.3\\
            &ResTune~\cite{liu2022residual} & 95.5 & 49.2 & 83.3 & 19.4 & 60.4 & 12.2 & 24.2 & 12.4 & 28.2 & 11.2\\
            &FRoST~\cite{Roy2022ClassincrementalNC} & \textbf{0.9} & 69.2 & 14.2 & 43.6 & 14.4 & 31.0 & 19.4 & 18.5 & 13.5 & 23.4 \\
            &CaSSLe~\cite{Fini2021SelfSupervisedMA} & 11.3 & 78.5 & 25.3 & 61.7 & 14.1 & 42.3 & 14.6 & 22.3 & 13.8 & 24.1\\
            
            \cline{2-12}
            
            &\ours & 8.2 & \textbf{85.4} & 15.6 & \textbf{63.7} & 9.2 & \textbf{53.3} & 13.7 & \textbf{28.9} & 3.1 & \textbf{25.2} \\
            &\ourspp & 7.6 & \textbf{91.7} & \textbf{12.3} & \textbf{67.7} & \textbf{1.6} & \textbf{56.5} & \textbf{0.6} & \textbf{41.1} & \textbf{2.7} & \textbf{26.1} \\
            
            \bottomrule
        \end{tabular}
    \end{center}
    \vspace{-1.0cm}
\end{table}

Tab.~\ref{tab:expt_sota} compares our proposed \ours and \ourspp with the adapted methods. ResTune underperforms in the \cincd setting due to its reliance on task-id information. FRoST exhibits strong ability to prevent forgetting on all data sets and sequences by segregating the \textit{not-forgetting} objective between the feature extractor and classifier. The adapted CIL methods capably discover new classes leveraging PTM knowledge. For two-step split sequences, these methods generally outperform \cincd adaptations by maintaining a balance between old and new classes. However, on five-step split sequences, the advantage of CIL-based methods over \cincd-based methods is not evident anymore, because CIL-based methods tend to forget tasks at the initial steps more when dealing with long sequences, as widely studied in CIL literature. EwC achieves better discovery accuracy by applying its forgetting prevention component directly to the model parameters using Fisher information matrix, while LwF~\cite{li2017learning} faces slow-fast learning interference issues. DER's performance suffers due to unstable self-supervised trajectories. CaSSLe is notably proficient in incremental discovery, attributed to its effective distillation mechanisms. Without \textit{bells} and \textit{whistles}, our \ours and \ourspp models consistently outperform adapted methods across datasets and sequences. While FRoST gives lower forgetting in some two-step split cases, our \ourspp, by improving the capacity for class-discrimination across all tasks, achieves lower forgetting in most five-step split cases.

\noindent\textbf{Generalizability Analysis.} Our proposed approach offers a versatile framework to convert related methods into effective \cincd solutions.
\begin{wrapfigure}{R}{0.49\textwidth}
    \vspace{-0.7cm}
    \centering
    \includegraphics[width=\linewidth]{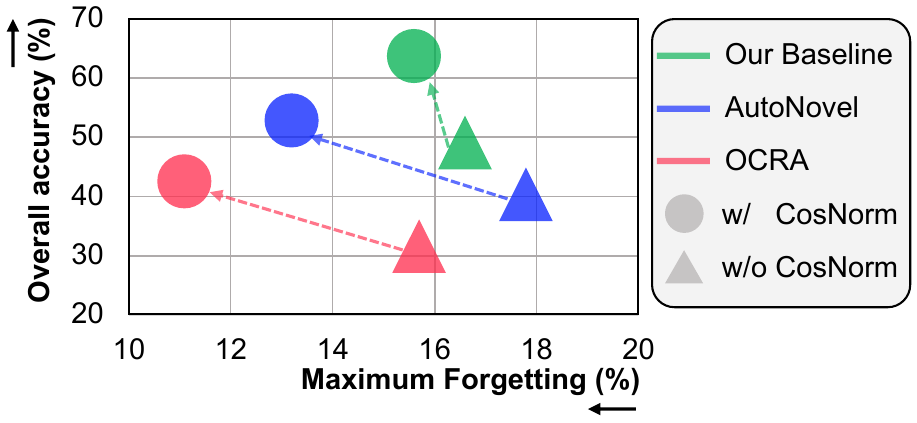}
    \vspace{-0.7cm}
    \caption{Generalizability analysis. Results are reported on the five-step split of C100 with DINO-ViT-B/16.}
    \label{fig:expt_wn_analysis}
    \vspace{-0.7cm}
\end{wrapfigure}
In Fig.~\ref{fig:expt_wn_analysis}, we equip two such methods, \ranks~\cite{han2020automatically} and \ocra~\cite{Cao2021OpenWorldSL}, with our proposed 
components (frozen PTM and \compcn). The results emphasize the pivotal role of \compcn in forming a task-agnostic classifier. Our findings reveal that, by removing \compcn, the converted methods suffer from significant forgetting due to non-uniformly scaled weight vectors, resulting in a decrease in overall discovery accuracy. This echoes the importance of \compcn in aligning the magnitude of the classifiers learned at each step to the same scale in \cincd scenarios.  Instead, with using \compcn, PTMs can be effectively leveraged to develop strong methods for the problem of \cincd.

\vspace{-3mm}
\section{Conclusion}
In this work we address the practical yet challenging task of Class-incremental Novel Class Discovery (\cincd). First, we highlight that the use of self-supervised pre-trained models (PTMs) can achieve better or comparable performance to models trained with labelled data in \ncd. Building upon this observation, we propose to forego the need for expensive labelled data by leveraging PTMs for \cincd. Second, we introduce two simple yet strong baselines that comprise of frozen PTM, \compcnlong and \compfrlong. Notably, our proposed baselines demonstrate significant improvements over the state-of-the-art methods across five datasets. We hope our work can provide a new, promising avenue towards effective \cincd.

\subsubsection{Acknowledgements}
E.R. is supported by MUR PNRR project FAIR - Future AI Research (PE00000013), funded by NextGenerationEU and EU projects SPRING (No. 871245) and ELIAS (No. 01120237). M.L. is supported by the PRIN project LEGO-AI (Prot. 2020TA3K9N). This work was carried out in the Vision and Learning joint laboratory of FBK and UNITN.

%%%%%%%%% APENDIX

%%%%%%%%% REFERENCE

%%%%%%%%%%% STD REF

\begin{thebibliography}{8}




\bibitem{Arthur2007kmeansTA} Arthur, D., Vassilvitskii, S.: k-means++: the advantages of careful seeding. In: ACM-SIAM Symposium on Discrete Algorithms (2007)

\bibitem{Boschini2022TransferWF} Boschini, M., Bonicelli, L., Porrello, A., Bellitto, G., Pennisi, M., Palazzo, S., Spampinato, C., Calderara, S.: Transfer without forgetting. In: ECCV (2022)

\bibitem{buzzega2020dark} Buzzega, P., Boschini, M., Porrello, A., Abati, D., Calderara, S.: Dark experience for general continual learning: a strong, simple baseline. In: NeurIPS (2020)

\bibitem{Cao2021OpenWorldSL} Cao, K., Brbic, M., Leskovec, J.: Open-world semi-supervised learning. In: ArXiv (2021)

\bibitem{Caron2020UnsupervisedLO} Caron, M., Misra, I., Mairal, J., Goyal, P., Bojanowski, P., Joulin, A.: Unsupervised learning of visual features by contrasting cluster assignments. In: NeurIPS (2020)

\bibitem{Caron2021EmergingPI} Caron, M., Touvron, H., Misra, I., J’egou, H., Mairal, J., Bojanowski, P., Joulin, A.: Emerging properties in self-supervised vision transformers. In: ICCV (2021)

\bibitem{Chang2017DeepAI} Chang, J., Wang, L., Meng, G., Xiang, S., Pan, C.: Deep adaptive image clustering. ICCV (2017)

\bibitem{Chen2020ASF} Chen, T., Kornblith, S., Norouzi, M., Hinton, G.E.: A simple framework for contrastive learning of visual representations. In: ArXiv (2020)

\bibitem{Cuturi2013SinkhornDL} Cuturi, M.: Sinkhorn distances: Lightspeed computation of optimal transport. In: NeurIPS (2013)

\bibitem{Deng2009ImageNetAL} Deng, J., Dong, W., Socher, R., Li, L.J., Li, K., Fei-Fei, L.: Imagenet: A large-scale hierarchical image database. In: CVPR (2009)

\bibitem{Dizaji2017DeepCV} Dizaji, K.G., Herandi, A., Deng, C., Cai, W.T., Huang, H.: Deep clustering via joint convolutional autoencoder embedding and relative entropy minimization. In: ICCV (2017)

\bibitem{Dosovitskiy2020AnII} Dosovitskiy, A., Beyer, L., Kolesnikov, A., Weissenborn, D., Zhai, X., Unterthiner, T., Dehghani, M., Minderer, M., Heigold, G., Gelly, S., Uszkoreit, J., Houlsby, N.: An image is worth 16x16 words: Transformers for image recognition at scale. In: ArXiv (2020)

\bibitem{Fei2022XConLW} Fei, Y., Zhao, Z., Yang, S.X., Zhao, B.: Xcon: Learning with experts for fine-grained category discovery. In: BMVC (2022)

\bibitem{Fini2021SelfSupervisedMA} Fini, E., Costa, V., Alameda-Pineda, X., Ricci, E., Karteek, A., Mairal, J.: Self-supervised models are continual learners. In: CVPR (2022)

\bibitem{fini2021unified} Fini, E., Sangineto, E., Lathuilière, S., Zhong, Z., Nabi, M., Ricci, E.: A unified objective for novel class discovery. In: ICCV (2021)

\bibitem{fomenko2022learning} Fomenko, V., Elezi, I., Ramanan, D., Leal-Taixé, L., Osep, A.: Learning to discover and detect objects. In: NeurIPS (2022)

\bibitem{French1999Catastrophic} French, R.: Catastrophic forgetting in connectionist networks. Trends in cognitive sciences (1999)

\bibitem{Han2019LearningTD} Han, K., Vedaldi, A., Zisserman, A.: Learning to discover novel visual categories via deep transfer clustering. In: ICCV (2019)

\bibitem{han2020automatically} Han, K., Rebuffi, S.A., Ehrhardt, S., Vedaldi, A., Zisserman, A.: Automatically discovering and learning new visual categories with ranking statistics. In: ICLR (2020)

\bibitem{He2015DeepRL} He, K., Zhang, X., Ren, S., Sun, J.: Deep residual learning for image recognition. In: CVPR (2015)

\bibitem{Hou2019LearningAU} Hou, S., Pan, X., Loy, C.C., Wang, Z., Lin, D.: Learning a unified classifier incrementally via rebalancing. In: CVPR (2019)

\bibitem{Hsu2017LearningTC} Hsu, Y.C., Lv, Z., Kira, Z.: Learning to cluster in order to transfer across domains and tasks. In: ArXiv (2017)

\bibitem{Hsu2019MulticlassCW} Hsu, Y.C., Lv, Z., Schlosser, J., Odom, P., Kira, Z.: Multi-class classification without multi-class labels. In: ArXiv (2019)

\bibitem{Jain2008DataC5} Jain, A.K.: Data clustering: 50 years beyond k-means. In: PRL (2008)

\bibitem{Jia2021JointRL} Jia, X., Han, K., Zhu, Y., Green, B.: Joint representation learning and novel category discovery on single- and multi-modal data. In: ICCV (2021)

\bibitem{Joseph2022NovelCD} Joseph, K.J., Paul, S., Aggarwal, G., Biswas, S., Rai, P., Han, K., Balasubramanian, V.N.: Novel class discovery without forgetting. In: ECCV (2022)

\bibitem{kirkpatrick2017overcoming} Kirkpatrick, J., Pascanu, R., Rabinowitz, N., Veness, J., Desjardins, G., Rusu, A.,
Milan, K., Quan, J., Ramalho, T., Grabska-Barwinska, A., Hassabis, D., Clopath, C., Kumaran, D., Hadsell, R.: Overcoming catastrophic forgetting in neural networks. Proceedings of the National Academy of Sciences (2016)

\bibitem{krizhevsky2009learning} Krizhevsky, A., Hinton, G., et al.: Learning multiple layers of features from tiny images (2009)

\bibitem{le2015tiny} Le, Y., Yang, X.: Tiny imagenet visual recognition challenge. CS 231N (2015)

\bibitem{li2017learning} Li, Z., Hoiem, D.: Learning without forgetting. In: TPAMI (2017)

\bibitem{lin2022continual} Lin, Z., Wang, Y., Lin, H.: Continual contrastive learning for image classification. In: 2022 IEEE International Conference on Multimedia and Expo (ICME) (2022)

\bibitem{liu2022residual} Liu, Y., Tuytelaars, T.: Residual tuning: Toward novel category discovery without labels. In: TNNLS (2022)

\bibitem{Luo2017CosineNU} Luo, C., Zhan, J., Wang, L., Yang, Q.: Cosine normalization: Using cosine similarity instead of dot product in neural networks. In: ArXiv (2017)

\bibitem{madaan2022representational} Madaan, D., Yoon, J., Li, Y., Liu, Y., Hwang, S.J.: Representational continuity for unsupervised continual learning. In: ICLR (2022), \url{https://openreview.net/forum?id=9Hrka5PA7LW}

\bibitem{masana2022class} Masana, M., Liu, X., Twardowski, B., Menta, M., Bagdanov, A.D., Van De Wei-jer, J.: Class-incremental learning: survey and performance evaluation on image classification. IEEE Transactions on Pattern Analysis and Machine Intelligence (2022)

\bibitem{Naseer2021IntriguingPO} Naseer, M., Ranasinghe, K., Khan, S.H., Hayat, M., Khan, F.S., Yang, M.H.: Intriguing properties of vision transformers. In: NeurIPS (2021)

\bibitem{Radford2021LearningTV} Radford, A., Kim, J.W., Hallacy, C., Ramesh, A., Goh, G., Agarwal, S., Sastry, G., Askell, A., Mishkin, P., Clark, J., Krueger, G., Sutskever, I.: Learning transferable visual models from natural language supervision. In: ICML (2021)

\bibitem{Rebuffi2016iCaRLIC} Rebuffi, S.A., Kolesnikov, A., Sperl, G., Lampert, C.H.: icarl: Incremental classifier and representation learning. In: CVPR (2016)

\bibitem{riz2023novel} Riz, L., Saltori, C., Ricci, E., Poiesi, F.: Novel class discovery for 3d point cloud semantic segmentation. In: CVPR (2023)

\bibitem{Roy2022ClassincrementalNC} Roy, S., Liu, M., Zhong, Z., Sebe, N., Ricci, E.: Class-incremental novel class discovery. In: ArXiv (2022)

\bibitem{Tan2019TheHC} Tan, K.C., Liu, Y., Ambrose, B.A., Tulig, M.C., Belongie, S.J.: The herbarium challenge 2019 dataset. In: ArXiv (2019)

\bibitem{troisemaine2023novel} Troisemaine, C., Lemaire, V., Gosselin, S., Reiffers-Masson, A., Flocon-Cholet, J., Vaton, S.: Novel class discovery: an introduction and key concepts. In: arXiv (2023)

\bibitem{Vaze2022GeneralizedCD} Vaze, S., Han, K., Vedaldi, A., Zisserman, A.: Generalized category discovery. In: CVPR (2022)

\bibitem{wang2023comprehensive} Wang, L., Zhang, X., Su, H., Zhu, J.: A comprehensive survey of continual learning: Theory, method and application. In: arXiv (2023)

\bibitem{wah2011caltech} Welinder, P., Branson, S., Mita, T., Wah, C., Schroff, F., Belongie, S., Perona, P.: Caltech-ucsd birds 200 (2010)

\bibitem{wu2022class} Wu, T.Y., Swaminathan, G., Li, Z., Ravichandran, A., Vasconcelos, N., Bhotika, R., Soatto, S.: Class-incremental learning with strong pre-trained models. In: CVPR (2022)

\bibitem{Wu2019LargeSI} Wu, Y., Chen, Y., Wang, L., Ye, Y., Liu, Z., Guo, Y., Fu, Y.R.: Large scale incremental learning. In: CVPR (2019)

\bibitem{Xie2015UnsupervisedDE} Xie, J., Girshick, R.B., Farhadi, A.: Unsupervised deep embedding for clustering analysis. In: ArXiv (2015)

\bibitem{Yang2016TowardsKS} Yang, B., Fu, X., Sidiropoulos, N., Hong, M.: Towards k-means-friendly spaces: Simultaneous deep learning and clustering. In: ICML (2016)

\bibitem{Yang2016JointUL} Yang, J., Parikh, D., Batra, D.: Joint unsupervised learning of deep representations and image clusters. In: CVPR (2016)

\bibitem{Yang2022DivideAC} Yang, M., Zhu, Y., Yu, J., Wu, A., Deng, C.: Divide and conquer: Compositional experts for generalized novel class discovery. In: CVPR (2022)

\bibitem{Zhao2021NovelVC} Zhao, B., Han, K.: Novel visual category discovery with dual ranking statistics and mutual knowledge distillation. In: ArXiv (2021)

\bibitem{zhong2021neighborhood} Zhong, Z., Fini, E., Roy, S., Luo, Z., Ricci, E., Sebe, N.: Neighborhood contrastive learning for novel class discovery. In: CVPR (2021)

\bibitem{Zhong2020OpenMixRK} Zhong, Z., Zhu, L., Luo, Z., Li, S., Yang, Y., Sebe, N.: Openmix: Reviving known knowledge for discovering novel visual categories in an open world. In: CVPR (2020)


%%%%%%%%% SUPP only
\bibitem{boschini2022class} Boschini, M., Bonicelli, L., Buzzega, P., Porrello, A., Calderara, S.: Class-incremental continual learning into the extended der-verse. In: TPAMI (2022)

\bibitem{Gou2020KnowledgeDA} Gou, J., Yu, B., Maybank, S.J., Tao, D.: Knowledge distillation: A survey. In: IJCV (2020)

\bibitem{Madaan2021RepresentationalCF} adaan, D., Yoon, J., Li, Y., Liu, Y., Hwang, S.J.: Representational continuity for unsupervised continual learning. In: ICLR (2021)

\bibitem{grill2020bootstrap} Grill, J.B., Strub, F., Altché, F., Tallec, C., Richemond, P., Buchatskaya, E., Doersch, C., Avila Pires, B., Guo, Z., Gheshlaghi Azar, M., et al.: Bootstrap your own latent-a new approach to self-supervised learning. In: NeurIPS (2020)

\end{thebibliography}

%%%%%%%%%%% ICPR 2024 REF

% \end{thebibliography}

\clearpage
% \maketitlesupplementary
\section*{Appendix}
\appendix

%%%%%%%%% APENDIX
This supplementary material is organized as follows: In Sec.~\ref{sec:app_expt_setup} we provide details about the datasets and the splits. In Sec.~\ref{sec:app-impl-det} we elaborate the implementation details of our baselines and the adapted methods under the \cincdtitle (\cincd) setting. In Sec.~\ref{sec:app-disc-uil}, we additionally discuss the fundamental difference between \cincd and \uillong (\uil), and also report the comparison with a SOTA \uil method. Sec.~\ref{sec:app-disc-overlap} extends the discussion on the decision to start multi-step class-incremental novel class discovery from self-supervised pre-trained models. We report the additional experimental results in more detail in Sec.~\ref{sec:app-exp-results}.

\section{Datasets and Splits}
\label{sec:app_expt_setup}

\begin{table}[!b]
     \vspace{-0.9cm}
    \begin{center}
      \centering
     \captionof{table}{Two-step and five-step dataset splits for the \cincd experiments. The number of novel classes $\classes\tst$ and the number of unlabelled images $\numdatast$ in $\data\tst$ for each task $\task\tst$ are reported.}
       \begin{tabular}{l|rrrr|rrrrrrrrrr}
        \toprule
        
        \multirow{3}{*}{Splits} & \multicolumn{4}{c|}{Two-step} & \multicolumn{10}{c}{Five-step} \\
         & \multicolumn{2}{c}{$\task\tsone$} & \multicolumn{2}{c|}{$\task\tstwo$}   & \multicolumn{2}{c}{$\task\tsone$} & \multicolumn{2}{c}{$\task\tstwo$} & \multicolumn{2}{c}{$\task\tsthree$} & \multicolumn{2}{c}{$\task\tsfour$} & \multicolumn{2}{c}{$\task\tsfive$} \\
        
         & $\classes\tsone$ & $\vert\data\tsone\vert$ & $\classes\tstwo$ & $\vert\data\tstwo\vert$ &  $\classes\tsone$ & $\vert\data\tsone\vert$ & $\classes\tstwo$ & $\vert\data\tstwo\vert$ & $\classes\tsthree$ & $\vert\data\tsthree\vert$ & $\classes\tsfour$ & $\vert\data\tsfour\vert$ & $\classes\tsfive$ & $\vert\data\tsfive\vert$ \\
        \hline
        
        C10 & 5 & 25.0k & 5 & 25.0k & 2 & 10.0k & 2 & 10.0k & 2 & 10.0k & 2 & 10.0k & 2 & 10.0k \\
        C100 & 50 & 25.0k & 50 & 25.0k & 20 & 10.0k & 20 & 10.0k & 20 & 10.0k & 20 & 10.0k & 20 & 10.0k \\
        T200 & 100 & 50.0k & 100 & 50.0k & 40 & 20.0k & 40 & 20.0k & 40 & 20.0k & 40 & 20.0k & 40 & 20.0k \\
        B200 & 100 & 2.4k & 100 & 2.4k & 40 & 0.9k & 40 & 0.9k & 40 & 0.9k & 40 & 0.9k & 40 & 0.9k \\
        H683 & 342 & 14.5k & 341 & 16.3k & 137 & 6.3k & 137 & 5.4k & 137 & 6.1k & 137 & 6.8k & 135 & 6.3k \\
        \bottomrule
      \end{tabular} 
       % \vspace{-.2in}

       \vspace{-0.9cm}
       \label{tab:data_details}
    \end{center}
\end{table}

We conduct experiments on five datasets, which are: CIFAR-10 (C10), CIFAR-100 (C100), TinyImageNet-200 (T200), CUB-200 (B200) and Herbarium-683 (H683). The Tab.~\ref{tab:data_details} presents the detailed splits for the two adopted task sequences (two-step and five-step) on the five data sets~\cite{krizhevsky2009learning,le2015tiny,wah2011caltech,Tan2019TheHC}. For a task sequence of $T=2$, the total classes and their corresponding instances in the dataset are equally divided into two splits (\eg, for C100, 100 classes / 2 tasks = 50 novel classes per task). Similarly, for a task sequence of $T=5$, the same method is used to divide the classes and their corresponding instances into five splits (\eg, for C100, 100 classes / 5 tasks = 20 novel classes per task).

The experimental results on C10, C100, and T200 provide an indication of the performance of the studied \cincd methods in common image recognition tasks, while the results on B200 and H683 show their performance in fine-grained image recognition tasks. Moreover, the evaluation on H683 offers insights into the performance of the studied methods in long-tailed task sequences and also when the downstream dataset is quite different from the internet-scale images.

\section{Implementation Details.}
\label{sec:app-impl-det}
In this section, we present the implementation details of image pre-processing and data augmentation, our baselines, and the adapted methods for \cincd. In Sec.~\ref{sec:supp_dataaug}, the employed image pre-processing technique and data augmentation for the experiments are elaborated. Sec.~\ref{sec:supp_ours} offers an in-depth account of the training and inference processes for our \ours and \ourspp using Pytorch-like pseudo-code. Subsequently, Sec.~\ref{sec:supp_bounds} explores the development of the reference methods for the \cincd setting. Ultimately, the adaption specifics and hyperparameters for the compared methods originating from \incd and \il fields are described in Sec.~\ref{sec:supp_incd} and Sec.~\ref{sec:supp_il}, correspondingly.

In order to maintain equitable evaluation, all methods examined in this work use the same \vitbsixteen~\cite{Dosovitskiy2020AnII} backbone, as employed by our \ours and \ourspp.

\subsection{Image Pre-processing and Data Augmentation}
\label{sec:supp_dataaug}
In order to utilize the publicly accessible pre-trained DINO-\vitbsixteen, it is necessary to adjust the input images to a fixed resolution of 224 $\times$ 224. In accordance with~\cite{Vaze2022GeneralizedCD}, the input images are initially upsampled to a resolution of 224 $\times$ 224 / 0.875 employing trilinear interpolation, followed by a center-crop of the upsampled images to achieve a 224 $\times$ 224 resolution for all the experiments. Subsequent to the aforementioned pre-processing procedure, SimCLR-like~\cite{Chen2020ASF} stochastic augmentations are predominantly employed throughout the experiments for all the methods.

\subsection{Simple yet Strong Baselines for \cincd}
\label{sec:supp_ours}
In the present section, a thorough exposition of the training and inference procedures for both \ours and \ourspp is provided, accompanied by Pytorch-like pseudo-code, to effectively demonstrate the simplicity of our methods.

\begin{algorithm}[!h]
\caption{Pseudo-code of our \ours training for the discovery task $\task\tst$ in a PyTorch-like style.}
\label{alg:ours_train}
% \algcomment{\fontsize{7.2pt}{0em}\selectfont \texttt{bmm}: batch matrix multiplication; \texttt{mm}: matrix multiplication; \texttt{cat}: concatenation.
% %\vspace{-1.em}
% }
\definecolor{codeblue}{rgb}{0.25,0.5,0.8}
\lstset{
  backgroundcolor=\color{white},
  basicstyle=\fontsize{7.2pt}{7.2pt}\ttfamily\selectfont,
  columns=fullflexible,
  breaklines=true,
  captionpos=b,
  commentstyle=\fontsize{7.2pt}{7.2pt}\color{codeblue},
  keywordstyle=\fontsize{7.2pt}{7.2pt},
%  frame=tb,
}
\begin{lstlisting}[language=python]
# g: frozen ViT-B/16 encoder network initialized by DINO weights, output 768-dimensional embedding
# h_t: task-specific linear classifier with randomly initialized weights
# temp: temperature
# C_t: number of novel classes present in task t

for x in train_loader:    # load a minibatch x with N samples
    x1 = aug(x)     # randomly augmented view 1
    x2 = aug(x)     # randomly augmented view 2

    # normalize weights
    with torch.no_grad():
        # temporarily store the weight vectors: C_tx768
        w_temp = h_t.linear_layer.weight.data.clone()
        w_temp = normalize(w_temp, dim=1, p=2)
        h_t.linear_layer.weight.copy_(w_temp)

    # extract feature embeddings
    z1 = g.forward(x1)      # Nx768
    z2 = g.forward(x2)      # Nx768

    # output logits
    l1 = h_t.forward(z1)    # NxC_t
    l2 = h_t.forward(z2)    # NxC_t

    # generate pseudo labels
    y1 = sinkhorn(l1)       # NxC_t
    y2 = sinkhorn(l2)       # NxC_t

    # swap prediction problem of the two views
    # cross-entropy loss, Eq.1
    loss1 = CrossEntropyLoss(l1/temp, y2)
    loss2 = CrossEntropyLoss(l2/temp, y1)
    loss = loss1 + loss2

    # SGD update: task-specific classifier
    loss.backward()
    update(h_t.params)
\end{lstlisting}
\end{algorithm}

\noindent
\textbf{Discovery Training.} Algo.~\ref{alg:ours_train} presents the pseudo-code for the shared discovery training loop employed by our \ours and \ourspp. For each unlabelled sample $\vx$, we generate two views of $\vx$ by applying the stochastic transformation delineated in Sec.~\ref{sec:supp_dataaug}. Prior to forwarding the input to the model, we perform cosine normalization on the task-specific classifier $h\tst$ through L2 normalization of the weight matrix $\theta_{h\tst}$ (feature-level L2 normalization is performed in $h\tst$). Subsequently, the two views are sequentially input into the frozen feature extractor $g$ and classifier $h\tst$ to produce the output logits. To optimize the objective defined in Eq.1 for the \textit{swapped} prediction problem, the \sinkhorn~\cite{Caron2020UnsupervisedLO} algorithm is employed to generate the pseudo-labels for the two views as soft-targets that are sawpped. The temperature parameter is set at 0.1. A total of 200 epochs of training are conducted for the discovery of both \ours and \ourspp.

\noindent
\textbf{Task-agnostic Inference.} After the discovery step for task $\task\tst$, our \ours can execute task-agnostic inference by simply concatenating the newly learned task-specific classifier $h\tst$ with the previous unified classifier $h\tstotminus$ to form a new unified classifier $h\tstot$, as described in Algo.~\ref{alg:ours_inference}.

\begin{algorithm}[!h]
\caption{Pseudo-code of our \ours inference for the task sequence $\taskb=\{\task\tsone, \task\tstwo, \cdots, \task\tsend\}$ in a PyTorch-like style.}
\label{alg:ours_inference}
% \algcomment{\fontsize{7.2pt}{0em}\selectfont \texttt{bmm}: batch matrix multiplication; \texttt{mm}: matrix multiplication; \texttt{cat}: concatenation.
% %\vspace{-1.em}
% }
\definecolor{codeblue}{rgb}{0.25,0.5,0.8}
\lstset{
  backgroundcolor=\color{white},
  basicstyle=\fontsize{7.2pt}{7.2pt}\ttfamily\selectfont,
  columns=fullflexible,
  breaklines=true,
  captionpos=b,
  commentstyle=\fontsize{7.2pt}{7.2pt}\color{codeblue},
  keywordstyle=\fontsize{7.2pt}{7.2pt},
%  frame=tb,
}
\begin{lstlisting}[language=python]
# g: frozen ViT-B/16 encoder network initialized by DINO weights, output 768-dimensional embedding
# h_prev: unifeid classifier saved for the previous tasks (t=1, ..., t=t-1)
# h_t: newly learned task-specific classifier for task t
# h_tot: unified classifier for all the tasks seen so far (t=1, ..., t=t)
# C_tot: total number of novel classes discovered until task t.

# concatenate classifiers
h_tot = cat([h_prev, h_t], dim=0)     # C_totx768

# perform task-agnostic inference
for x in test_loader:    # load a minibatch x with N samples
    # extract feature embeddings
    z = g.forward(x)        # Nx768

    # output logits
    l = h_tot.forward(z)    # NxC_tot

    # take the cluster-id with maximum logit value as prediction
    prediction = max(l, dim=1)   # N
\end{lstlisting}
\end{algorithm}

\noindent
\textbf{\compfr Training.} As expounded in the primary manuscript, our \ourspp incorporates \textit{\compfrlong} (\compfr) training to jointly optimize the concatenated classifier $h\tstot$ further. Algo.~\ref{alg:ourspp_train} describes the \compfr training loop specifics. For each unlabelled sample $\vx$, the same stochastic transformation (see Sec.~\ref{sec:supp_dataaug}) is employed to generate two views of $\vx$. The \textit{cosine normalization} operation is applied to the unified classifier before forward propagation to maintain the weight vectors on the same scale. Subsequently, generative pseudo feature replay is utilized to replay an equal number of past feature embeddings from the preserved pseudo per-class prototype Gaussian distributions $\Mu$ as the current mini-batch size. The loss, as defined in Eq. 4 for past novel classes, is calculated using the output logits of the replayed embeddings from $h\tstot$. To also preserve the discriminative capability for current novel classes, knowledge is transferred from $h\tst$ to $h\tstot$ using the pseudo-labels generated by $h\tst$ for the two views. The loss, as defined in Eq. 5 for the current novel classes, is then computed using the output logits of the two views' embeddings from $h\tstot$ and the pseudo-labels. The ultimate \textit{past-current} objective (refer to Eq. 6) for \ourspp training is optimized by aggregating the two individual losses to update the parameters $\theta_{h\tstot}$ of the unified classifier $h\tstot$.

\begin{algorithm}[!h]
\caption{Pseudo-code of the \compfr training in \ourspp during task $\task\tsend$ in a PyTorch-like style.}
\label{alg:ourspp_train}
% \algcomment{\fontsize{7.2pt}{0em}\selectfont \texttt{bmm}: batch matrix multiplication; \texttt{mm}: matrix multiplication; \texttt{cat}: concatenation.
% %\vspace{-1.em}
% }
\definecolor{codeblue}{rgb}{0.25,0.5,0.8}
\lstset{
  backgroundcolor=\color{white},
  basicstyle=\fontsize{7.2pt}{7.2pt}\ttfamily\selectfont,
  columns=fullflexible,
  breaklines=true,
  captionpos=b,
  commentstyle=\fontsize{7.2pt}{7.2pt}\color{codeblue},
  keywordstyle=\fontsize{7.2pt}{7.2pt},
%  frame=tb,
}
\begin{lstlisting}[language=python]
# concatenate classifiers
h_tot = cat([h_prev, h_t], dim=0)     # C_totx768

# load a minibatch x with N samples
for x in train_loader:
    x1 = aug(x)     # randomly augmented view 1
    x2 = aug(x)     # randomly augmented view 2

    # normalize weights
    with torch.no_grad():
        # temporarily store the weight vectors: C_totx768
        w_temp = h_tot.linear_layer.weight.data.clone()
        w_temp = normalize(w_temp, dim=1, p=2)
        h_tot.linear_layer.weight.copy_(w_temp)

        
    # generatively replay saved prototypes fro past classes
    z_past, y_past = replay(M)  # Nx768, Nx1

    # output logits for past embeddings from unified classifier
    l_past = h_tot.forward(z_past)  # NxC_tot
    
    # cross-entropy loss for past classes, Eq.4
    loss_past = CrossEntropyLoss(l_past/temp, y_past)

    # extract feature embeddings
    z1 = g.forward(x1)      # Nx768
    z2 = g.forward(x2)      # Nx768

    # output logits
    l1 = h_T.forward(z1)    # NxC_t
    l2 = h_T.forward(z2)    # NxC_t

    # generate pseudo labels by using the task-specific classifier prediction
    y1 = max(l1, dim=1) + C_tot - C_t   # Nx1
    y2 = max(l2, dim=1) + C_tot - C_t   # Nx1

    # concatenate feature embeddings
    z_current = cat([z1, z2], dim=0)    # 2Nx768
    
    # concatenate pseudo labels
    y_current = cat([y1, y2], dim=0)    # 2Nx1

    # output logits for current embeddings from unified classifier
    l_now = h_tot.forward(z_current)    # 2NxC_tot
    
    # cross-entropy loss for current classes, Eq.5
    loss_current = CrossEntropyLoss(l_now/temp, y_current)
    
    # swap prediction problem of the two views
    # cross-entropy loss, Eq.6
    loss = l_past + loss_current

    # SGD update: task-specific classifier
    loss.backward()
    update(h_tot.params)
\end{lstlisting}
\vspace{+10mm}
\end{algorithm}

\subsection{Building Reference Methods}
\label{sec:supp_bounds}
Since there is no prior work has investigated \cincd setting, we build reference methods for the comparison in this work.

\noindent
\textbf{K-means~\cite{Arthur2007kmeansTA}.} We utilize the K-means algorithm to create a 'pseudo' \textit{lower-bound} reference. Specifically, we extract the 768-dimensional deep features $\vz \in \R^{768}$ of the given images using DINO-\vitbsixteen~\cite{Caron2021EmergingPI} as the feature extractor. Then, we perform K-means clustering on the $\vz$ extracted from the joint training datasets $\bigcup_{t=1}^{T} \data\tst$ to form $\bigcup_{t=1}^{T} \classes\tst$ semantic clusters at the end of a given task sequence. The maximum number of iterations is set to 300 for all the experiments. However, these \textit{lower-bound} results are only for reference as the K-means algorithm uses access to previous training data to form the clusters for task-agnostic evaluation and cannot accurately represent the minimum performance of \cincd.

\noindent
\textbf{Joint (frozen) and Joint (unfrozen).} In accordance with the supervised \cil practice~\cite{wang2023comprehensive}, we construct two \textit{upper-bound} reference methods. The first method, denoted as $\upperb$, performs joint training on the unified model $f\tstoend = h\tstoend \circ g$ of \ours after task-specific discovery training, utilizing all the training data $\bigcup_{t=1}^{T} \data\tst$ up to the current step. The second method, denoted as $\upperbpp$, further unfreezes the last transformer block during both the discovery and joint training of $\upperb$. Notably, $\upperbpp$ does not unfreeze the last block at the beginning of the training since we observe in experiments that saturating the classifier $h\tst$ first and then fine-tuning the last block of $g$ yields better performance.

\subsection{Adapting \incd Methods to \cincd}
\label{sec:supp_incd}

% In this section, we describe the implementation details of ResTune~\cite{liu2022residual} and FRoST~\cite{Roy2022ClassincrementalNC}, which are adapted to \cincd from \incd.
Since \cincd setting does not allow the use of any labelled data, we need to adapt the two compared \incd solutions, ResTune\footnote {\url{https://github.com/liuyudut/ResTune}}~\cite{liu2022residual} and FRoST\footnote {\url{https://github.com/OatmealLiu/class-iNCD}}~\cite{Roy2022ClassincrementalNC}, to work without the supervised pre-training on the labelled data. To accomplish this, we initialize the feature extractors $g(\cdot)$ of ResTune and FRoST with the same self-supervised pre-trained weights $\theta_{g}$ (DINO~\cite{Caron2021EmergingPI}), instead of using supervised pre-training on labelled data. This enables ResTune and FRoST to perform continuous novel class discovery under the \cincd setting with their own components to discover novel categoires and prevent forgetting.

\noindent
\textbf{ResTune} is an \incd solution that combines architecture-based and regularization-based \il techniques to prevent forgetting. ResTune grows a new block at each incremental step to learn new knowledge with a clustering objective~\cite{Xie2015UnsupervisedDE}, while adjusting the shared basic feature extractor for the new data under the regularization of a knowledge distillation objective~\cite{li2017learning}. The adapted ResTune in this work uses the first eleven transformer blocks of \vitbsixteen as the shared basic feature extractor, with only the last (11th) block unfrozen, while creating a new unfrozen transformer block branch initialized by DINO-weights to learn the residual feature at each step. The weight $\beta$ for the knowledge distillation objective is set to 1 for all the experiments, as in the original work.

\noindent
FRoST is a \cincd solution that combines regularization-based and rehearsal-based \il techniques to prevent forgetting, and it is based on ranking statistics~\cite{han2020automatically}. In this work, we strictly follow the configuration used in the original work~\cite{Roy2022ClassincrementalNC} for the hyperparameters. However, since there are no labels available in the \cincd setting, we adapt the supervised feature replay of FRoST to the unsupervised pseudo feature replay by using the same approach in \ourspp.

\subsection{Adapting \il Methods to \cincd}
\label{sec:supp_il}

% In this section, we describe the implementation details of EwC~\cite{kirkpatrick2017overcoming}, LwF~\cite{li2017learning}, and DER~\cite{buzzega2020dark}, which are adapted to \cincd from \il.
To evaluate the effectiveness of our proposed methods in preventing forgetting, we also compare their performance with that of traditional \il techniques. For this purpose, we adapt two regularization-based methods (EwC~\cite{kirkpatrick2017overcoming} and LwF~\cite{li2017learning}) and one rehearsal-based method (DER~\cite{buzzega2020dark}) to \cincd by using the publicly available \il framework codebase\footnote{\url{https://github.com/aimagelab/mammoth}}~\cite{buzzega2020dark,boschini2022class} in our experiments. However, unlike the \incd methods, these \il methods are originally designed for supervised settings and are not capable of discovering novel categories from unlabelled data. Therefore, we apply the same discovery strategy as our \ours to all the adapted \il methods. Specifically, we initialize the feature extractor $g$ with DINO~\cite{Caron2021EmergingPI} pre-trained weights and optimize the clustering objective defined in Eq.1 using the \sinkhorn cross-view pseudo-labelling algorithm~\cite{Caron2020UnsupervisedLO} to discover the novel classes contained in the given unlabelled data set $\data\tst$. Different from our \ours, we unfreeze the last transformer block of $g$ to adapt the model to the data present at each step in all the experiments. To prevent forgetting, we maintain the \il components in the original methods.

\noindent
\textbf{EwC} is a weight regularization \il method, which penalizes the model parameters selectively based on their importance for the past tasks using the calculated Fisher information matrix~\cite{kirkpatrick2017overcoming}. In the experiments, we set the hyperparameter $\lambda$ to 8000 to control the relative importance of past tasks compared to the new one, and the Fisher matrix fusion parameter $\alpha$ to 0.5.

\noindent
\textbf{LwF} is a function regularization \il solution that uses a knowledge distillation~\cite{Gou2020KnowledgeDA} objective function to prevent forgetting by constraining the current model output to not deviate too much from the old model~\cite{li2017learning}. In our experiments, we save the old model $f\tstotminus = h\tstotminus \circ g\tstminus$ to compute the LwF loss at each step $t$. The LwF loss weight $\lambda$, which determines the balance between the old and new tasks, is set to 1.0 for all experiments.

\noindent
\textbf{DER} is a rehearsal-based \il solution that involves storing a fixed-size buffer of old training samples with past model responses as proxies of old tasks to prevent forgetting~\cite{buzzega2020dark}. For our experiments, the adapted DER maintains a buffer of 500 old samples for each step, with the \textit{not-forgetting} loss weight $\alpha$ set to 0.5.

\section{Comparison with Unsupervised Incremental Learning Method}
\label{sec:app-disc-uil}
We wish to emphasize that the primary focus of our study is on incrementally discovering and grouping of novel classes rather than incremental representation learning. This distinguishes our setting from \uiltitle (\uil) task~\cite{Madaan2021RepresentationalCF,Fini2021SelfSupervisedMA}. {\uil has the following \textit{drawbacks} w.r.t our proposed \textbf{\cincd}: \textbf{i}) UIL deals with learning \textit{only} the backbone, whereas in \cincd one can learn both the backbone and the classifier; \textbf{ii}) UIL methods either need \textit{labelled} data to train a classifier, or require access to \textit{past training data} for k-NN classification, both of which are \textbf{\textit{not}} needed by \cincd methods.} Thus, we believe \cincd is \textbf{more general} and subsumes the UIL methods.

Even so, we adapted and compared with a SOTA UIL method, \textbf{CaSSLe}\footnote{\url{https://github.com/DonkeyShot21/cassle}}~\cite{Fini2021SelfSupervisedMA} (CVPR'22), on the two task splits of the five benchmarks with the same DINO-initialized \vitbsixteen as feature extractor $g$. At each step discovery step, CaSSLe first trains $g$ on $\data\tst$ to learn the representation with its self-supervised loss (BYOL~\cite{grill2020bootstrap}) and distillation loss. Being a \uil method, CaSSLe requires labelled samples to learn a classifier, which are not available in the \cincd setting. To learn the classifier \textit{unsupervisedly} for discovery, we equip CaSSLe with our self-labelling loss (Eq. 1) for \ncd and our CosNorm to make it task-agnostic for a fair comparison. The results is reported in Tab.~\ref{tab:expt_cassle}. We observe that our simpler baselines consistently outperform CaSSLe both in accuracy ($\accuracy$) and forgetting ($\forgetting$) metrics. The worse performance by CaSSLe suggests that although it adopts distillation mechanisms to map the current representations at each step back to the previous steps, learning a classifier along with the backbone unsupervisedly is quite intricate, which we try to resolve with our proposed Baselines.

\begin{table}[!t]
    % \tablestyle{1.0pt}{0.9}
    \small
    % \vspace{-0.7cm}
\caption{Comparison with the adapted state-of-the-art \uil method on two task splits of C10, C100, T200, B200, and H683 under \cincd setting. Overall accuracy and maximum forgetting are reported. All methods use DINO-\vitbsixteen as feature encoder.}
    \label{tab:expt_cassle}
    % \vspace{-0.3cm}
    \begin{center}
        \begin{tabular}{cl|cc|cc|cc|cc|cc}
            \toprule
            
            & Datasets 
            & \multicolumn{2}{c|}{C10} 
            & \multicolumn{2}{c|}{C100} 
            & \multicolumn{2}{c|}{T200}
            & \multicolumn{2}{c|}{B200}
            & \multicolumn{2}{c}{H683} \\

            & Methods 
            & $\forgetting\downarrow$ 
            & $\accuracy\uparrow$ 
            & $\forgetting\downarrow$ 
            & $\accuracy\uparrow$ 
            & $\forgetting\downarrow$ 
            & $\accuracy\uparrow$ 
            & $\forgetting\downarrow$ 
            & $\accuracy\uparrow$ 
            & $\forgetting\downarrow$ 
            & $\accuracy\uparrow$ \\
            \hline
    
            \multirow{3}{*}{\rotatebox[origin=c]{90}{\textbf{2-step}}}
            &CaSSLe~\cite{Fini2021SelfSupervisedMA} & 9.1 & 87.3 & 10.3 & 53.7 & 6.9 & 36.5 & 4.8 & 26.8 & 10.9 & 25.3 \\
            
            \cline{2-12}
            
            &\ours & 8.5 & \textbf{89.2} & 6.7 & \textbf{60.3} & 4.0 & \textbf{54.6} & 4.1 & \textbf{28.7} & 7.9 & \textbf{25.7} \\
            &\ourspp & 4.5 & \textbf{90.9} & 6.6 & \textbf{61.4} & \textbf{0.2} & \textbf{55.1} & 4.2 & \textbf{36.9} & \textbf{6.0} & \textbf{27.5}\\
            
            \toprule
            \multirow{3}{*}{\rotatebox[origin=c]{90}{\textbf{5-step}}}
            &CaSSLe~\cite{Fini2021SelfSupervisedMA} & 11.3 & 78.5 & 25.3 & 61.7 & 14.1 & 42.3 & 14.6 & 22.3 & 13.8 & 24.1\\
            
            \cline{2-12}
            
            &\ours & 8.2 & \textbf{85.4} & 15.6 & \textbf{63.7} & 9.2 & \textbf{53.3} & 13.7 & \textbf{28.9} & 3.1 & \textbf{25.2} \\
            &\ourspp & 7.6 & \textbf{91.7} & \textbf{12.3} & \textbf{67.7} & \textbf{1.6} & \textbf{56.5} & \textbf{0.6} & \textbf{41.1} & \textbf{2.7} & \textbf{26.1} \\
            
            \bottomrule
        \end{tabular}
    \end{center}
    % \vspace{-1.0cm}
\end{table}

\section{Is it fair to use Self-supervised Pre-trained Models (PTMs) for \cincdlong?}
\label{sec:app-disc-overlap}
We believe the usage of self-supervised pre-trained models (PTMs) is \textbf{justified} due to several reasons: \textbf{i}) the PTM (\textbf{DINO}) was pre-trained \textit{without} labels, thus complying with the one of the assumptions made in NCD; \textbf{ii}) our proposal to use PTM can be viewed analogous to the Generalized Category Discovery (\textbf{GCD})~\cite{Vaze2022GeneralizedCD}, where the unlabelled samples can come from both previously \textit{seen} and \textit{unseen} classes; and \textbf{iii}) in real-world scenarios of clustering with a large pool of data, one would normally start from a \textit{generic} pre-trained model, \textit{without} having any knowledge about the pre-training classes. Thus, the core idea -- of \textbf{leveraging prior knowledge} to better cluster unlabelled data -- remains unchanged. Guided with these motivations, we start from a self-supervised PTM and show with extensive experiments that a simple Baseline is more adept at \cincd than many SOTA methods proposed in related areas. In addition, to fairly exam our methods, we validate the proposed baselines on \textbf{diverse} and \textbf{balanced} datasets. Two out of the five datasets, CUB-200 and Herb-19, \textbf{do not} significantly overlap with ImageNet. In detail, only 2 classes in CUB-200 \textit{exactly} overlap with ImageNet. Herb-19 is disjoint in its \textbf{entirety}, which is evident from the lowest performance among all the datasets.

\section{Detailed Experimental Results}
\label{sec:app-exp-results}

\begin{figure}[!b]
\vspace{-4mm}
\begin{center}
    \includegraphics[width=0.95\linewidth]{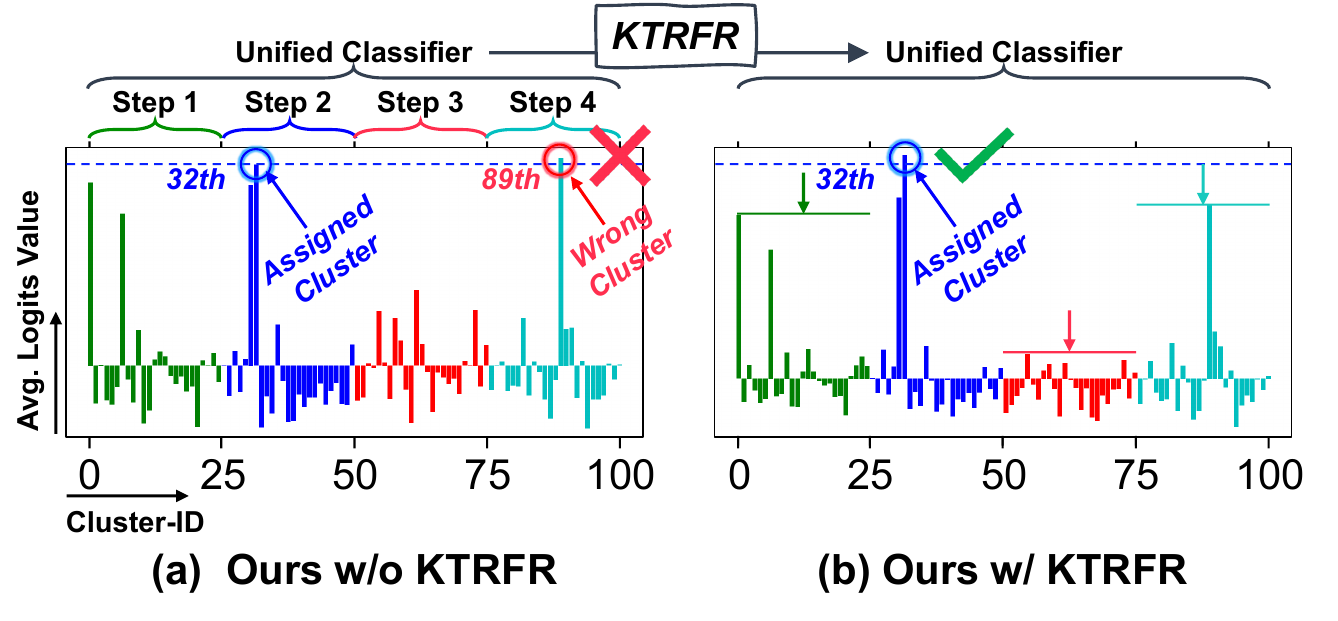}
\end{center}
% \vspace{-1mm}
\caption{Average output logits of our (a) \ours and (b) \ourspp for instances sampled from the 47th-class of C100. Results are evaluated on the four-step split at the end of the discovery task sequence.}
\label{fig:expt_featreplay}
\vspace{-6mm}
\end{figure}

\noindent \textbf{Qualititative Analysis of \compfrtitle (\compfr).} To better understand the benefit of \compfr, we present a qualitative analysis in Fig.~\ref{fig:expt_featreplay}. We show the average logit values obtained from the unified classifier $h\tstoend$, with and without using \compfr, for the test images of the $\texttt{47th}$ category in the four-step split C100. The plot reveals that in the absence of \compfr (see Fig.~\ref{fig:expt_featreplay}a), the logit corresponding to the incorrect cluster ($\texttt{89th}$, red-circled, discovered at the fourth task) exhibits a higher value compared to the correct cluster ($\texttt{32th}$, blue-circled, discovered at the second task), which corresponds to the $\texttt{47th}$ category. However, by incorporating \compfr (see Fig.~\ref{fig:expt_featreplay}b), the logits associated with the incorrect clusters become less active, as well as, the logit value for the correct cluster exhibits an increase, leading to better performance. This further demonstrates the effectiveness of \compfr in discriminating classes among all tasks.

\begin{figure}[!h]
\begin{center}
\includegraphics[width=0.9\linewidth]{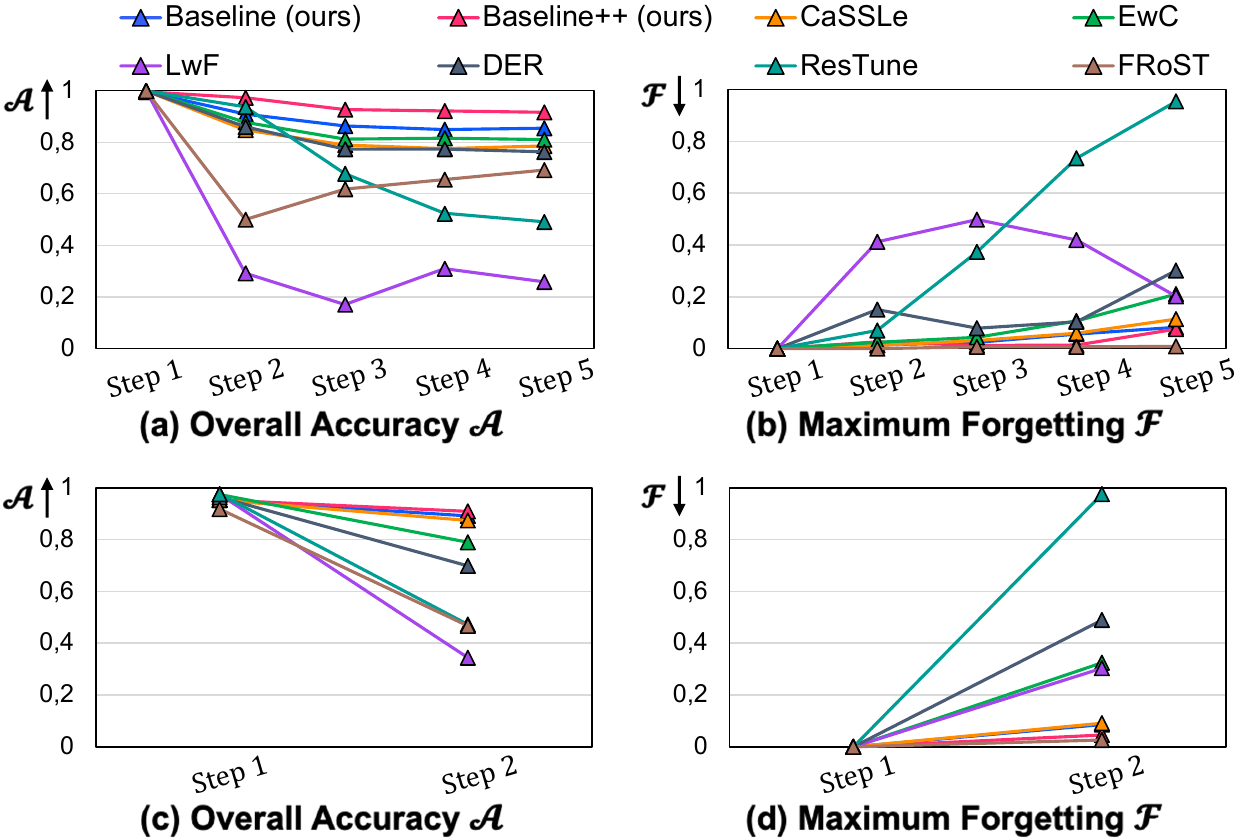}
\end{center}
\vspace{-1mm}
\caption{Comparison of our baseline methods with the adapted state-of-the-art methods (EwC, LwF, DER, ResTune, FRoST, CaSSLe) on \textbf{C10} under the \cincd setting. \textbf{Top (a, b)}: five-step split. \textbf{Bottom (c, d)}: two-step split. The overall accuracy and maximum forgetting are reported.}
\label{fig:supp_sota_c10}
\vspace{-5mm}
\end{figure}

\begin{figure}[!h]
\begin{center}
\includegraphics[width=0.9\linewidth]{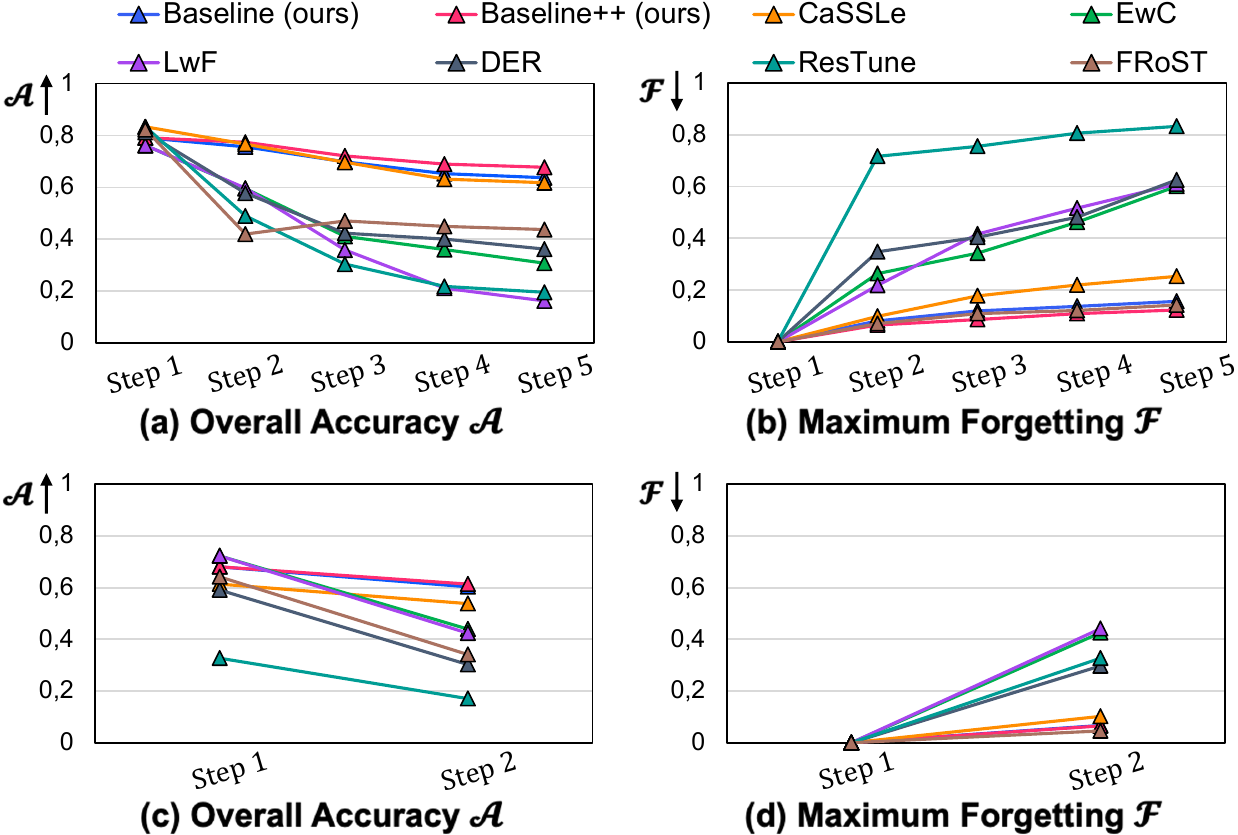}
\end{center}
\vspace{-1mm}
\caption{Comparison of our baseline methods with the adapted state-of-the-art methods (EwC, LwF, DER, ResTune, FRoST, CaSSLe) on \textbf{C100} under the \cincd setting. \textbf{Top (a, b)}: five-step split. \textbf{Bottom (c, d)}: two-step split. The overall accuracy and maximum forgetting are reported.}
\label{fig:supp_sota_c100}
\vspace{-5mm}
\end{figure}

\begin{figure}[!tbh]
\begin{center}
\includegraphics[width=0.9\linewidth]{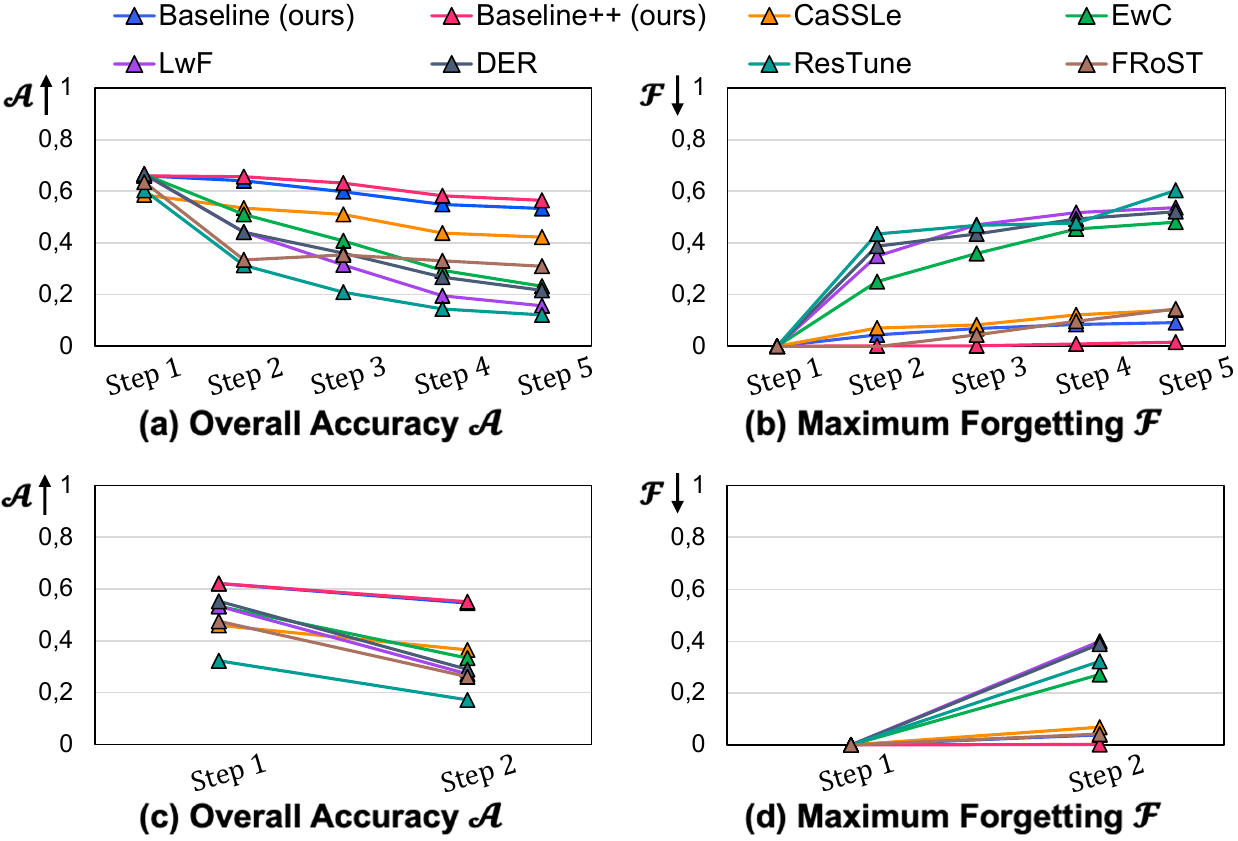}
\end{center}
\vspace{-1mm}
\caption{Comparison of our baseline methods with the adapted state-of-the-art methods (EwC, LwF, DER, ResTune, FRoST, CaSSLe) on \textbf{T200} under the \cincd setting. \textbf{Top (a, b)}: five-step split. \textbf{Bottom (c, d)}: two-step split. The overall accuracy and maximum forgetting are reported.}
\label{fig:supp_sota_t200}
\vspace{-4mm}
\end{figure}

\begin{figure}[!tbh]
\begin{center}
\includegraphics[width=0.9\linewidth]{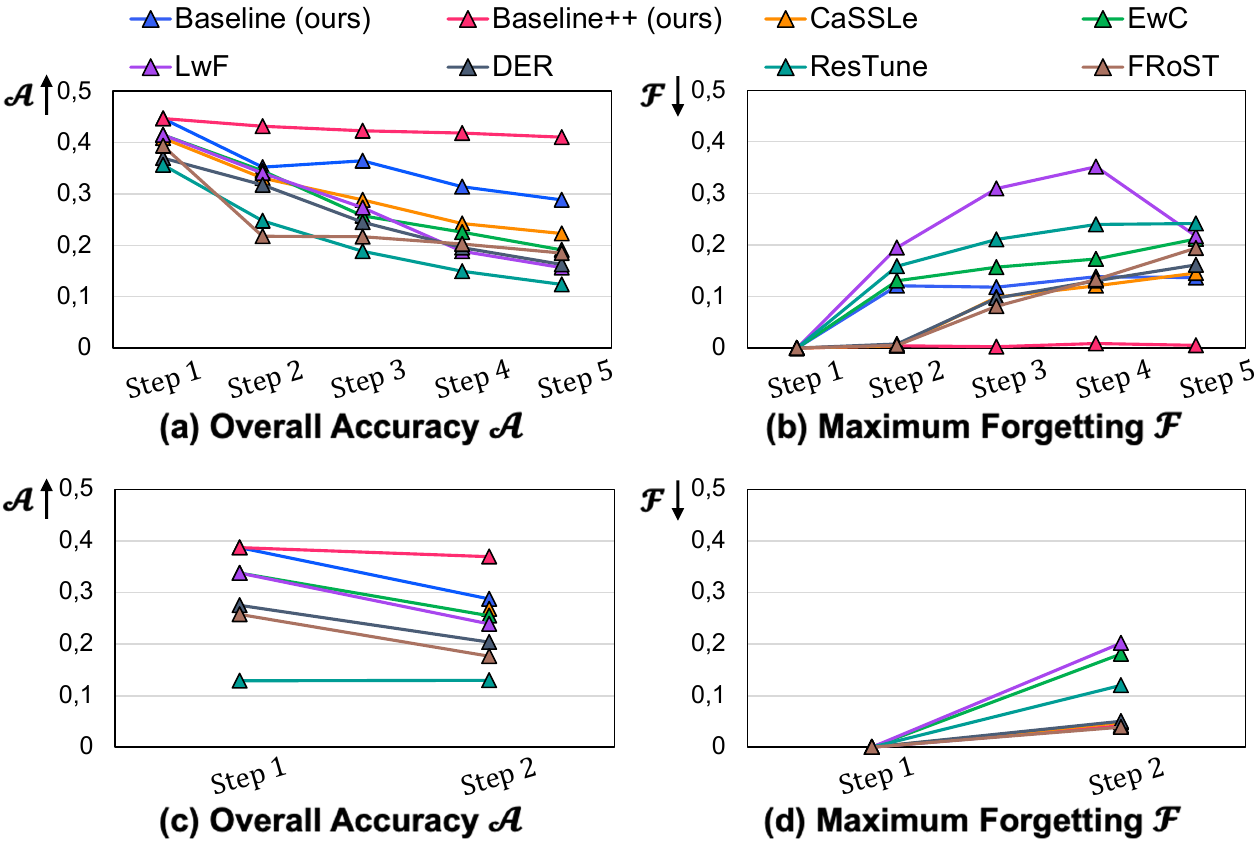}
\end{center}
\vspace{-1mm}
\caption{Comparison of our baseline methods with the adapted state-of-the-art methods (EwC, LwF, DER, ResTune, FRoST, CaSSLe) on \textbf{B200} under the \cincd setting. \textbf{Top (a, b)}: five-step split. \textbf{Bottom (c, d)}: two-step split. The overall accuracy and maximum forgetting are reported.}
\label{fig:supp_sota_b200}
\vspace{-4mm}
\end{figure}

\begin{figure}[!tbh]
\begin{center}
\includegraphics[width=0.9\linewidth]{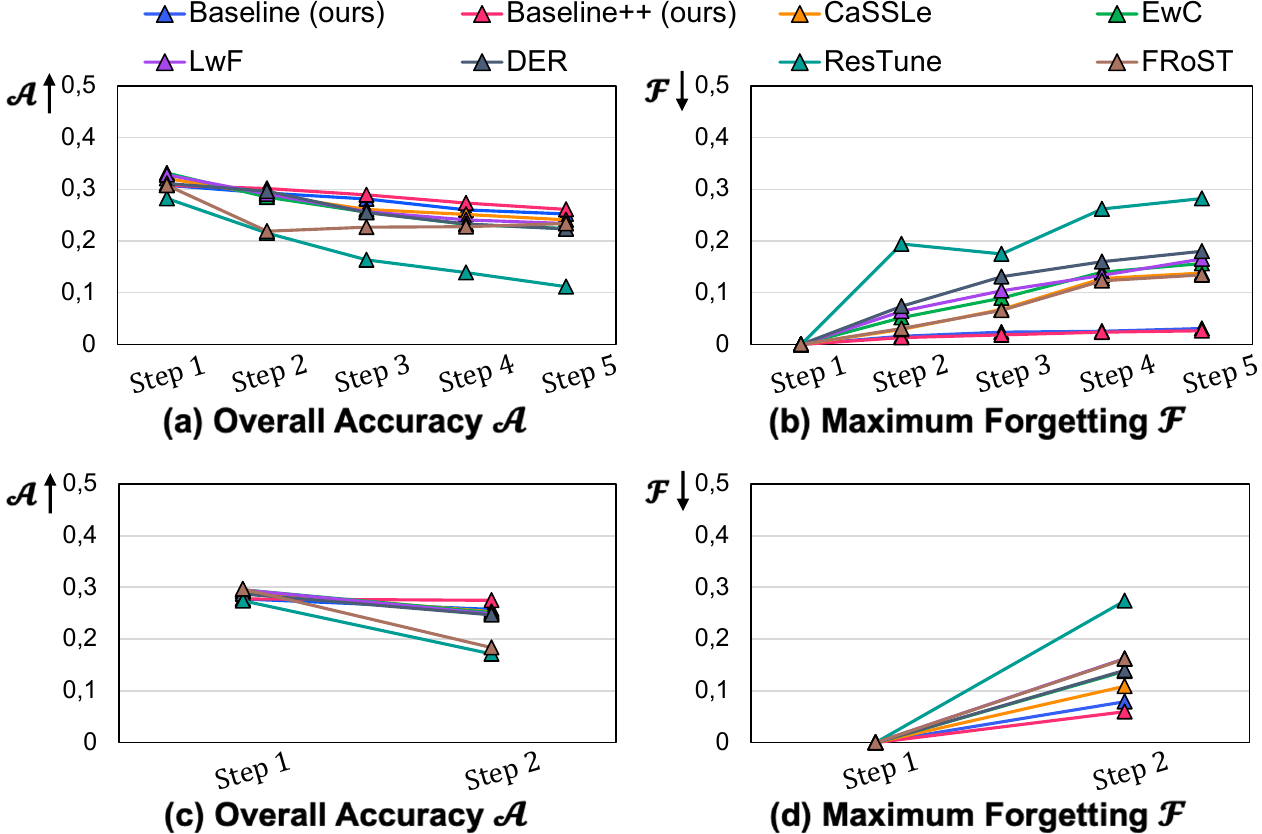}
\end{center}
\vspace{-1mm}
\caption{Comparison of our baseline methods with the adapted state-of-the-art methods (EwC, LwF, DER, ResTune, FRoST, CaSSLe) on \textbf{H683} under the \cincd setting. \textbf{Top (a, b)}: five-step split. \textbf{Bottom (c, d)}: two-step split. The overall accuracy and maximum forgetting are reported.}
\label{fig:supp_sota_h683}
\vspace{-4mm}
\end{figure}

\noindent \textbf{Per-step Comparison with the State-of-the-art Methods.} In this section, we provide comprehensive per-step comparative results of our \ours and \ourspp, juxtaposed with the adapted state-of-the-art methods on two task splits (two-step and five-step) of CIFAR-10 (C10)~\cite{krizhevsky2009learning}, CIFAR-100 (C100)~\cite{krizhevsky2009learning}, TinyImageNet-200 (T200)~\cite{le2015tiny}, CUB-200 (B200)~\cite{wah2011caltech} and Herbarium-683 (H683)~\cite{Tan2019TheHC} under \cincd setting in Fig.~\ref{fig:supp_sota_c10}, Fig.~\ref{fig:supp_sota_c100}, Fig.~\ref{fig:supp_sota_t200}, Fig.~\ref{fig:supp_sota_b200}, Fig.~\ref{fig:supp_sota_h683}, respectively. We report both the overall accuracy and maximum forgetting for each step, employing a task-agnostic evaluation.

As depicted in the reported figures, the overall accuracy exhibits a decline as the task sequence progresses, whereas the maximum forgetting for the novel classes discovered during the first step experiences an increase, attributable to the \textit{\forget} issue~\cite{wang2023comprehensive}. In the context of longer task sequences (five-step split, as observed in the top half of the figures), the \textit{forgetting} issue is exacerbated due to more frequent model updates.

During the first discovery task, the majority of adapted methods that unfreeze the final transformer block attain higher accuracy in most cases due to their adaptation to the current data. However, commencing from the second step, our \ours and \ourspp consistently surpass all compared methods across all datasets and splits in terms of overall accuracy. Although FRoST~\cite{Roy2022ClassincrementalNC} exhibits a better ability to mitigate forgetting for novel classes discovered in the first step in certain cases, our baselines demonstrate a more balanced performance between the past and current novel classes. The consistent experimental results from the five compared datasets and two task splitting strategies reiterate the preeminence of our \ours and \ourspp for the \cincd task. A straightforward combination of existing \il components and \ncd solutions proves insufficient for the \cincd task. While the two very recent \incd works (ResTune~\cite{liu2022residual} and FRoST~\cite{Roy2022ClassincrementalNC}) were designed for such unsupervised incremental scenarios, they fail to achieve satisfactory performance when the restrictive assumption of possessing a rich labelled base classes is relaxed. Conversely, our proposed baseline methods operate without the need for labelled base classes; nevertheless, utilizing rich labelled data to pre-supervise the self-supervised PTM can also be employed in our \ours and \ourspp to enhance single-step \ncd performance if such labelled data is accessible.

Lastly, upon comparing the top half (a and b) with the bottom half (c and d) of all the presented figures, it becomes evident that the accuracy/forgetting disparities between \ours and \ourspp widen as the task sequence lengthens. This observation underscores the significance and efficacy of the \compfr training employed by \ourspp in enhancing the class-discrimination capability across tasks.

\clearpage

\end{document}